%% file: main.tex
\documentclass[10pt,journal,compsoc]{IEEEtran}
\usepackage[nocompress]{cite}
\usepackage{graphicx}
\usepackage{caption} 
\usepackage{subcaption}
\usepackage{amsmath}
\usepackage{amsfonts}
\usepackage{amssymb}
\usepackage{cancel}
\usepackage{algorithm}
\usepackage{algpseudocode}
\usepackage{multirow} %
\usepackage{multicol}
\usepackage{adjustbox}
\usepackage{booktabs} 
\usepackage{bm}
\usepackage{adjustbox}
\usepackage[dvipsnames]{xcolor}
\usepackage{hyperref} 
\definecolor{mydarkblue}{rgb}{0,0.08,0.45}
\definecolor{myfavblue}{rgb}{0.1176, 0.392, 1.0}
\hypersetup{
colorlinks=true,
 linkcolor=mydarkblue,
 citecolor=mydarkblue,
 filecolor=mydarkblue,
  urlcolor=mydarkblue}
    
\begin{document}

\title{A Deterministic Approximation to Neural SDEs}

\author{Andreas~Look,
        Melih~Kandemir,
        Barbara~Rakitsch,
        and~Jan~Peters,~\IEEEmembership{Fellow,~IEEE}
\IEEEcompsocitemizethanks{\IEEEcompsocthanksitem A. Look and B. Rakitsch  are with the Bosch Center for Artificial Intelligence, Renningen, Germany.\protect\\
E-mail: \{andreas.look, barbara.rakitsch\}@bosch.com
\IEEEcompsocthanksitem M. Kandemir was with the  Bosch Center for Artificial Intelligence. He is now with the University of Southern Denmark, Odense, Denmark. \protect\\
E-mail: kandemir@imada.sdu.dk. 
\IEEEcompsocthanksitem J. Peters is with Intelligent Autonomous Systems Group, TechnicalUniversity Darmstadt, Darmstadt, Germany, and also with the Max Planck Institute for Intelligent Systems, 72076 Tübingen, Germany. \protect\\
E-mail: peters@ias.informatik.tu-darmstadt.de.}
\thanks{}}
\markboth{}%
{Shell \MakeLowercase{\textit{et al.}}: Bare Demo of IEEEtran.cls for Computer Society Journals}

\IEEEtitleabstractindextext{%

\begin{abstract}
Neural Stochastic Differential Equations (NSDEs) model the drift and diffusion functions of a stochastic process as neural networks. While NSDEs are known to make accurate predictions, their uncertainty quantification properties have been remained unexplored so far. We report the empirical finding that obtaining well-calibrated uncertainty estimations from NSDEs is computationally prohibitive. As a remedy, we develop a computationally affordable deterministic scheme which accurately approximates the transition kernel, when dynamics is governed by a NSDE. Our method introduces a bidimensional moment matching algorithm: vertical along the neural net layers and horizontal along the time direction, which benefits from an original combination of effective approximations. Our deterministic approximation of the transition kernel is applicable  to both training and prediction. We observe in multiple experiments that the uncertainty calibration quality of our method can be matched by Monte Carlo sampling only after introducing high computational cost. Thanks to the numerical stability of deterministic training, our method also improves prediction accuracy.
\end{abstract}

}

\maketitle
\IEEEdisplaynontitleabstractindextext
\IEEEpeerreviewmaketitle

\input{sections/introduction.tex}
\input{sections/nsde.tex}

\input{sections/bmm.tex}
\input{sections/layers.tex}
\input{sections/numerical_experiments.tex}
\input{sections/experiments.tex}
\input{sections/related_work.tex}
\input{sections/conclusion.tex}

\bibliographystyle{IEEEtran}
\bibliography{main}
\input{sections/biography.tex}
\end{document}

%% file: sections/introduction.tex
\IEEEraisesectionheading{\section{Introduction}\label{sec:introduction}}
\IEEEPARstart{A}{ccompanying} 
 time series predictions with calibrated uncertainty scores is a challenging problem. The main difficulty is that uncertainty assessments for individual time points propagate, causing local errors to impair the predictions on the whole sequence. While calibrated prediction is well-studied in the regression setting \cite{Platt99probabilisticoutputs, calibration_via_aux}, the same problem is a relatively new challenge for predictors with feedback loops. The few prior work is restricted to post-hoc calibration of deterministic recurrent neural nets \cite{accurate_uncertainties, calibrated_regression}. 

We report the first study on the uncertainty quantification characteristics of {\it Neural Stochastic Differential Equations} (NSDEs) \cite{nsde_raginsky, adjoint_sde}. A Neural SDE models the continuous dynamics of an environment with a drift neural net governing the deterministic component of a vector field and a diffusion neural net governing the instantaneous distortions. Neural SDEs have a large potential to provide an attractive tool to the machine learning community due to their strong theoretical links to \textit{Recurrent Neural Nets} (RNNs), \textit{Neural ODEs} (NODEs) \cite{neural_ode}, and \textit{Gaussian Processes} (GPs) \cite{applied_sde, diffgp}.  While multiple studies have observed NSDEs to bring encouraging success in prediction accuracy, none has thus far investigated their performance in uncertainty quantification. Yet, an essential benefit of modeling stochasticity is to account for uncertainty in a reliable way. 

\begin{figure*}[t]
	\centering
	\includegraphics[width=2\columnwidth]{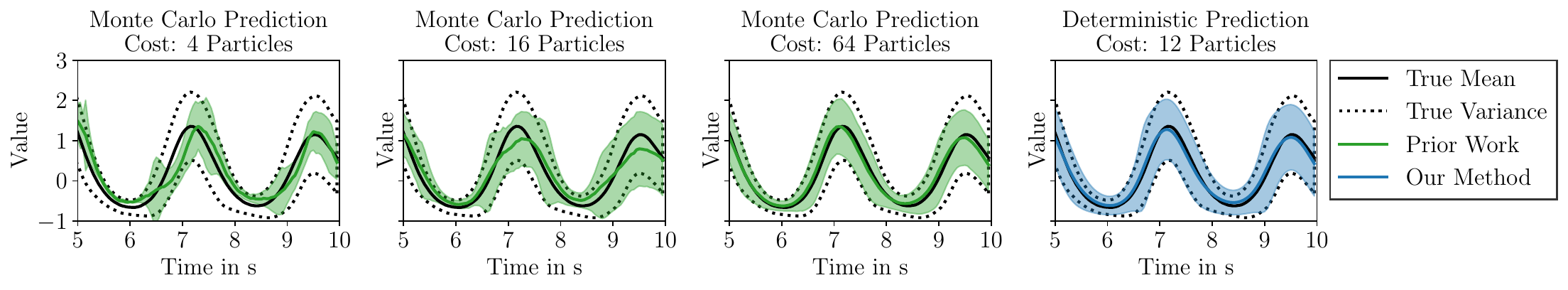}
	\caption{Our deterministic approximation  provides well-calibrated uncertainty scores with a computational cost equal to 12 particles. Reaching a comparable level of calibration by Monte Carlo sampling demands at least 64 particles.
	\label{fig:fig1}}
\end{figure*}

We focus on a central observation: when dynamics is governed by a NSDE, accurate approximation of the transition kernel, which describes a density for a given point in time, with Monte Carlo sampling requires a prohibitively large sample set, i.e. computation time. 
We introduce an original method for deterministic approximation of the transition kernel that can deliver well-calibrated prediction uncertainties at significantly lower computational cost than Monte Carlo sampling by the virtue of
\begin{itemize}
    \item [i)] performing \textit{Bidimensional Moment Matching (BMM)} to approximate the intractable expectation and covariance integrals: horizontally across time and vertically across the layers of drift and diffusion neural nets, 
    \item [ii)] using Steins's lemma to simplify calculation of covariances while matching moments across time, and
    \item [iii)] approximating the expected Jacobian of a neural net accurately while matching moments across layers.
\end{itemize}
 As visualized for a toy case in Fig. \ref{fig:fig1}, NSDE prediction with Monte Carlo sampling requires a large sample set to catch up with the calibration level of our deterministic method.

 We investigate the numerical properties of  BMM in Sec. \ref{sec:numerical_properties} and provide benchmarks in Sec. \ref{sec:experiments} on various applications.

 
%% file: sections/nsde.tex
\section{Neural Stochastic Differential Equations}
\label{sec:nsde}

We are concerned with the model family that describes the dynamics of a $D-$dimensional stochastic process $x(t)$ as a non-linear time-invariant SDE
\begin{align}
d x(t) = f_{\theta}(x(t)) d t  +L_{\phi}(x(t)) d w(t).
\label{eq:nsde}
\end{align}
Above, $f_{\theta}(x(t)): \mathbb{R}^D \rightarrow \mathbb{R}^D$ is the drift function governing the deterministic component of the SDE, which is modeled as a neural net with an arbitrary architecture. Similarly, the diffusion function $L_{\phi}(x(t)): \mathbb{R}^{D } \rightarrow  \mathbb{R}^{D\times D}$ is another neural net, which models the stochasticity of the system. 
\textcolor{black}{We assume that both $f_\theta$ and $L_\phi$ are neural nets, which are parameterized by $\theta$ and $\phi$, with varying number of hidden layers and activation functions that ensure Lipschitz-continuity $\forall x(t), x'(t): |f_{\theta}(x(t))-f_{\theta}(x'(t))|+|L_{\phi}(x(t))-L_{\phi}(x'(t))| \leq c |x(t)-x'(t)|$ for some constant $c$ and linear growth $\forall x(t): |f_\theta(x(t))| + |L_\phi(x(t))| \leq d (1 + |x(t)|)$ for some constant $d$. 
The two requirements ensure the existence of the solution to Eq. \ref{eq:nsde} \cite{oksendal}. 
}

Further, $dt$ is the time increment and $w(t)$ is a  $D-$dimensional standard Wiener process that injects noise into the dynamics following a normal distribution with variance proportional to the time increment. 
In order to make a prediction at time point $t_{n+1}$, an initial value $x(t_n)$ at time point $t_n$ needs to be provided and the below system to be solved
\begin{align}
    x(t_{n+1})\! =\! x(t_n) + \int_{t_n}^{t_{n+1}}\! f_{\theta}(x(t)) d t  + \int_{t_n}^{t_{n+1}}\! L_{\phi}(x(t)) d w(t).
\label{eq:nsde_solution}
\end{align}
The integral involving the drift function is defined as a Lebesgue integral and the second integral involving the diffusion is defined as an It\^{o} integral \cite{oksendal}. The randomness induced at every infinitesimal time step by $dw(t)$ makes the solution $x(t_{n+1})$ a random variable that follows a probability distribution \textcolor{black}{$p(x(t_{n+1})|x(t_n), \theta, \phi)$. For the sake of brevity we omit in the following the dependence on 
\textcolor{black}{$\theta$ and $\phi$.}}

The transition kernel $p(x(t_{n+1})|x(t_n))$ can be accessed by solving the \textit{Fokker–Planck-Kolmogorov} (FPK) equation, which is a potentially high-dimensional \textcolor{black}{partial differential equation} with an often intractable solution. As the transition kernel is necessary for \textcolor{black}{likelihood-based} parameter inference schemes of SDEs  as well as for uncertainty quantification, various approximations of $p(x(t_{n+1})|x(t_n))$ have been proposed, e.g. sampling based approaches \cite{brandt, petersen_95, ll_inference_mcmc}, methods based on approximate solutions to the FPK equation \cite{HURN1999373, Jensen18}, or deterministic approximations based on an assumed density \cite{saerkkae_inference}. 
Prior work in the context of NSDEs \cite{nsde_raginsky, dbnn, adjoint_sde} commonly approximates the transition kernel via Monte Carlo methods. 
As we find out that sampling noise impairs the predictive calibration (see Fig. \ref{fig:fig1}), we derive a novel deterministic approximation of the transition kernel, which efficiently exploits the layered structure of neural networks.

%% file: sections/bmm.tex
\section{Bidimensional Moment Matching}
\label{sec:bmm}
In this section, we derive a novel deterministic approximation to the transition kernel $p(x(t_{n+1})|x(t_n))$, when dynamics is governed by a NSDE as defined in Eq. \ref{eq:nsde}.
We craft our solution in four steps: 
(i) discretizing the NSDE in Eq. \ref{eq:euler_maruyama}, 
(ii) approximating the process distribution at every discretization point as a normal density in Eq. \ref{eq:mom-match}, (iii) analytically marginalizing out the Wiener process noise from moment matching update rules in Eqs. \ref{eq:moment-matching-mean} and \ref{eq:moment-matching-covariance}, (iv) approximating the intractable terms in the moment calculations in Sec. \ref{subsec:drift_moments}, \ref{subsec:diff_moments}, and \ref{subsec:cross_cov}. 

Instead of solving the FPK equation, we may obtain an approximation to the transition kernel $p(x(t_{n+1})|x(t_n))$ by firstly solving  for the next state $x(t_{n+1})$ as in Eq. \ref{eq:nsde_solution}. 
Since the Wiener process injects randomness to any arbitrarily small time interval, the solution of any NSDE with drift and diffusion networks with at least one hidden layer is analytically intractable. 
As a numerical approximation  to Eq. \ref{eq:nsde_solution}, we adopt the {\it Euler-Maruyama} (EM) method due to its computational efficiency. 
For notational clarity, we assume an evenly spaced discretization, though the time step size $\Delta t > 0$ can be chosen dynamically if desired. 
We discretize the interval $t_{n+1}-t_n$ into $K$ steps such that $t_{n+1}=t_n + K\Delta t$.
Without loss of generality, we refer to the discretized version of $x(t_n)$ as $x_0$, respectively to $x(t_{n+1})$ as $x_K$.
The EM method follows the update rule
\begin{align}
&x_{k+1} := x_k +  f_{\theta}(x_k)\Delta t+L_{\phi}(x_k) w_k, \label{eq:euler_maruyama}
\end{align}
where $ w_k \sim \mathcal{N}(0, \Delta t)$. 
This solution amounts to the below approximation of the transition kernel
\begin{align}
    p(x_{k+1}|x_k) := \mathcal{N}  (x_{k+1}  | m_{k+1}(x_k), S_{k+1}(x_k) ), \label{eq:em_normal}
\end{align}
where $m_{k+1}(x_k) := x_k +  f_{\theta}(x_k)\Delta t$ and $S_{k+1}(x_k) := {L}_{\phi}{L}_{\phi}^T(x_{k})  \Delta t$ are the mean and covariance of a normal density.
Given the above approximation of the transition kernel, we may express the $k$-step transition density for a given $x_0$ as a series of nested integrals
\begin{equation}
    p(x_{k}|x_0)  = \int  p(x_{k}|x_{k-1})p(x_{k-1}|x_{0})   d x_{k-1}.
    \label{eq:recursion}
\end{equation} 
The above expression needs to be solved recursively  for $k \in \{ 1, \ldots, K\}$ for a given $x_0$ with $p(x_1 | x_0)$ given by Eq. \ref{eq:em_normal}.
Since the EM method is consistent \cite{applied_sde} we obtain $p(x_{K}|x_0) = p(x(t_{n+1})|x(t_n))$ as $\Delta t \rightarrow 0$.  
Prior work on NSDEs \cite{dbnn, adjoint_sde} evaluates this intractable recurrence relation via MC integration. 
After sampling multiple trajectories from the discretized NSDE as defined in Eq. \ref{eq:euler_maruyama}, the dependence on the previous time step  in Eq. \ref{eq:recursion} can be  marginalized out, resembling the transition kernel approximation in the simulated maximum likelihood method \cite{brandt, petersen_95}.

\subsection{Assumed Process Density}
\label{sec:gaussian_assumption}
As the solution to the nested integrals, which describes the $k$-step transition density of a discretized NSDE (Eq. \ref{eq:recursion}), is intractable for non-trivial architectures, we approximate the transition kernel at every step $k$ by a normal density 
\begin{align}
p(x_{k}|x_0) \approx \mathcal{N}(x_{k}|\mu_{k},\Sigma_{k}), \label{eq:mom-match}
\end{align}
with mean $\mu_{k}$ and covariance $\Sigma_{k}$. This approximation simplifies the problem to calculating the first two moments of the transition kernel.  Plugging the {\it Assumed Density (AD)} in Eq. \ref{eq:mom-match} into the recurrence relation for estimation of $p(x_{k}|x_0)$, as defined in Eq. \ref{eq:recursion}, amounts to approximating the transition kernel at every time point by matching moments progressively in time direction. We refer to this chain of operations as {\it Horizontal Moment Matching } (HMM).

Calculating $\mu_{k}$ and $\Sigma_{k}$ does not appear to be a simpler problem at the first sight than solving Eq. \ref{eq:recursion}. However, it is possible to obtain a more pleasant expression by reparametrizing   $p(x_{k}|x_0)$ as
\begin{align}
~~\zeta_{k-1} \sim N(\zeta_{k-1}|0,I), ~~~x_{k-1} \sim \mathcal{N}(x_{k-1}|\mu_{k-1}, \Sigma_{k-1}), \\
    x_{k} := x_{k-1} + f_{\theta}(x_{k-1})\Delta t + L_{\phi} (x_{k-1})\sqrt{\Delta t} \zeta_{k-1}, \nonumber 
\end{align}
where $I$ is the identity matrix with appropriate dimensionality. We arrive at the following  view of the first moment of $p(x_{k}|x_0)$ using the law of the unconscious statistician 
\begin{align}
  \mu_{k} &=   \mathbb{E} [x_{k-1} + f_\theta(x_{k-1})\Delta t + L_{\phi} (x_{k-1})\sqrt{\Delta t} \zeta_{k-1}] \label{eq:moment-matching-mean}\\
  &=  \mu_{k-1} +  \mathbb{E} [f_\theta(x_{k-1})\Delta t]. \nonumber
\end{align}
In order to derive a tractable expression for the variance $\Sigma_{k} = \mathbb{E}[x_{k}x_{k}^T] - \mathbb{E}[x_{k}]\mathbb{E}[x_{k}]^T$, 
we first evaluate 
\begin{flalign}
 \mathbb{E}[x_{k}x_{k}^T]   
  =\mathbb{E}\big[&(x_{k-1}\!+\!{f}_\theta(x_{k-1})\Delta t)(x_{k-1}\!+\!{f}_\theta(x_{k-1})\Delta t)^T\big]+&&\nonumber\\
   \mathbb{E}\big[&{L}_{\phi}{L}_{\phi}^T(x_{k-1}) \big] \Delta t,&&
\end{flalign}
since $\mathbb{E}[\zeta_{k-1}]=0$ and $\mathbb{E}[\zeta_{k-1} \zeta_{k-1}^T]= I$.
Using the bilinearity of the covariance operator, we obtain 
\begin{align}
    \Sigma_{k} = 
    &\Sigma_{k-1}\!+ \mathrm{Cov}[f_\theta( x_{k-1})]  \Delta t^2\!+\mathrm{Cov}[ f_\theta( x_{k-1}), x_{k-1}] \Delta t +   \nonumber\\ 
     &\mathrm{Cov}[ f_\theta( x_{k-1}), x_{k-1}]^T \Delta t\! + \mathbb{E}[{L}_{\phi}{L}_{\phi}^T(x_{k-1})] \Delta t,
     \label{eq:moment-matching-covariance}
\end{align}
where $\mathrm{Cov}[f_\theta( x_{k-1}), x_{k-1}]$ denotes the cross-covariance between the random vectors in the arguments. 
Eq. \ref{eq:moment-matching-mean} and \ref{eq:moment-matching-covariance} have no closed form solution for neural nets and require numerical approximation.
Moment matching solutions along similar lines have been developed earlier for SDEs \cite{saerkkae_inference, saerkka_smoothing}, which rely on standard numerical integration schemes. 
In contrast to prior work we  develop in the following sections an integration scheme, which efficiently uses the layered structure of neural nets, resulting in a more accurate and faster method.

\subsection{Computing the Drift Network Moments} 
\label{subsec:drift_moments}
After applying the HMM scheme, the terms $\mathbb{E}[f_\theta( x_{k})]$ and $\mathrm{Cov}[f_\theta(x_{k})]$ amount to the first two moments of a random variable obtained by propagating $x_{k} \sim \mathcal{N}(\mu_{k}, \Sigma_{k})$ through the neural net 
$f_\theta(x_{k}) := u_L ( u_{L-1} ( \ldots u_2( u_1(x_{k}) ) \ldots ) )$,
composed of a chain of $L$ simple functions (layers), typically an alternation of affine transformations and nonlinear activations. Calculating the moments of $f_\theta(x_{k})$ is analytically intractable due to the nonlinear activations. We approximate this computation by another round of moment matching, this time by propagating the input noise through the neural net. Denote the feature map at layer $l$ at time step ${k}$ as $h_{k}^l := u_l ( u_{l-1} ( \ldots  u_2 ( u_1(x_{k}) ) \ldots ) )$, which is a random variable due to $x_{k}$ and is related recursively to the feature map of the previous layer as $h_{k}^l = u_l (h_{k}^{l-1})$. Denoting $h_{k}^0 := x_{k}$, we approximate the distributions on layers recursively as 
\begin{align}
    h_{k}^l = u_l(h_{k}^{l-1}) \approx  \widetilde{h}_{k}^l \sim \mathcal{N}(a_{k}^l, B_{k}^l), \label{eq:vmm}
\end{align}
where $a_{k}^l:=\mathbb{E}[u_l( \widetilde{h}_{k}^{l-1})]$ and $B_{k}^l := \mathrm{Cov}[u_l(\widetilde{h}_{k}^{l-1})]$. We refer to applying this approximation throughout all neural net layers  as {\it Vertical Moment Matching } (VMM). As an outcome of VMM, we get $\mathbb{E} [ f_\theta(x_{k}) ] \approx a_{k}^L$ and $\mathrm{Cov}[f_\theta( x_{k})] \approx B_{k}^L$. We provide output moments $a_{k}^l$ and $B_{k}^l$ for commonly used layers in Sec. \ref{sec:layers}. 

A similar approach has been applied earlier to \textit{Bayesian Neural Nets} (BNN) for random weights in various contexts such as expectation propagation \cite{pbp, ghosh}, deterministic variational inference \cite{wu2018deterministic}, and evidential deep learning \cite{bedl}. To our knowledge, no prior work has applied this approach to propagating input uncertainty through a deterministic network in the dynamics modeling context.

\subsection{Computing the Diffusion Network Moments} 
\label{subsec:diff_moments}
We assume the diffusion matrix $L_{\phi}(x_k)$ \textcolor{black}{to be diagonal with positive entries only}, though our method generalizes trivially to a full diffusion matrix. Overloading the notation for the sake of brevity, we denote an $L-$layer neural net assigned to its diagonal entries as $L_{\phi}(x_k) :=  v_L ( v_{L-1} ( \ldots v_2( v_1(x_k) ) \ldots ) )$ with feature maps defined recursively as $e_k^l := v_l (e_k^{l-1})$. Following VMM, we pass the moments through the network by approximating the random (due to noisy $x_k$) feature map $e_k^l$ by a normal distribution  $\widetilde{e}_k^l \sim \mathcal{N}(c_k^l, D_k^l)$ applying the moment matching rules to Eq. \ref{eq:vmm} literally on $c_k^l$ and $D_k^l$ and get as output
\begin{equation}
\mathbb{E}[{L}_{\phi}{L}_{\phi}^T(x_{k})] \approx (D_k^L + (c_k^L)(c_k^L)^T) \odot I   := D_k,
\label{eq:diff_central_moment}
\end{equation}
with $\odot$ denoting the Hadamard product. The second central moment $D_k$ is a diagonal matrix due to the restriction of a vector-valued output of ${L}_{\phi}(x_k)$. 

\subsection{Computing the cross-covariance} 
\label{subsec:cross_cov}
The term $\mathrm{Cov}[f_\theta(x_k), x_k]$ corresponds to the cross-covariance between the input $x_k$, which is a random variable, and its transformation with the drift function $f_\theta(x_k)$.  Due to the same reasons as the mean and covariance of $f_\theta(x_k)$, this cross-covariance term cannot be analytically calculated except for trivial drift functions. However, cross-covariance is not provided as a direct outcome of VMM. As being neither a symmetric nor a positive semi-definite matrix,  inaccurate approximation of cross-covariance may impair numerical stability. 
Applying Stein's lemma \cite{steins_lemma} 
\begin{align}
    \mathrm{Cov}[x_k, f_\theta(x_k)] = \mathrm{Cov}[x_k] \mathbb{E}[\nabla_{x_k} f_\theta(x_k)] \label{eq:stein}
\end{align}
for the first time in the context of matching moments of a neural net, we obtain a form that is easier to approximate. The covariance $\mathrm{Cov}[x_k]$ is provided from the previous time step as $\Sigma_k$, but the expected gradient $\mathbb{E}[\nabla_{x_k} f_\theta(x_k)]$ needs to be explicitly calculated. In standard BNN inference, where the source of uncertainty are the weights, we have the interchangeability property for the gradients with respect to the weight distribution parameters $\xi$ as $\nabla_\xi \mathbb{E}_{q(\theta|\xi)}[ f_\theta(x_k)] = \mathbb{E}_{q(\theta|\xi)}[\nabla_\xi f_\theta(x_k)]$.
This trick is not applicable to our case as the gradient is with respect to $x_k$. In other words, $\mathbb{E}[\nabla_{x_k} f_\theta(x_k)]$ is not equal to $\nabla_{x_k} \mathbb{E}[ f_\theta(x_k)]$, which would otherwise allow us to simply use $\nabla \mu_k$. 

Applying the chain rule, the expectation of the derivative of a neural net with respect to a random input reads
\begin{align}
   \mathbb{E} [\nabla_{x_k} f_\theta(x_k) ]  =\mathbb{E} [    \nabla_{x_k} u_1 (x_k) \ldots  \nabla_{h_k^{L-1}} u_L (h_k^{L-1})], \label{eq:exp_grad}
\end{align}
which is also analytically intractable. We facilitate computation by making the assumption that the mutual information between nonlinear feature maps of different layers is small  
\begin{align}
\int p(h_k^l, h_k^{l'})  \log \Bigg \{\frac{p(h_k^l, h_k^{l'})}{ p(h_k^l) p(h_k^{l'})} \Bigg \} dh_k^l dh_k^{l'}  \approx 0,    
\end{align}
for all pairs $(l,l')$ with $l \neq l'$ and $1 \leq l, l', \leq L$.  Applying this assumption of decoupled activations on Eq. \ref{eq:exp_grad}, we get
\begin{align}
  \mathbb{E} [\nabla_{x_k} f_\theta(x_k) ]  
  \approx \prod_{l=0}^{L-1} \mathbb{E}_{{h}_k^l}  [\nabla_{{h}_k^l} u_{l+1}({h}_k^l)]. \label{eq:indep_activations}
\end{align}

\begin{figure}[t]
	\begin{subfigure}[t]{.3\columnwidth}
		\centering\includegraphics[height=.9\columnwidth]{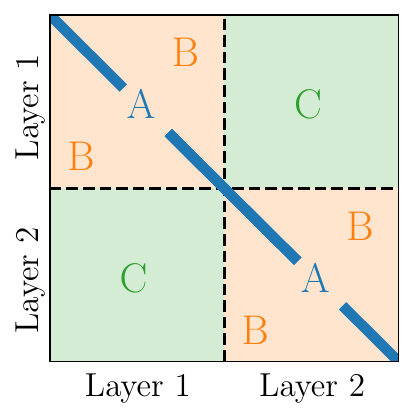}
		\caption{Legend.}
	\end{subfigure}%
	\begin{subfigure}[t]{.3\columnwidth}
	\includegraphics[height=.9\columnwidth]{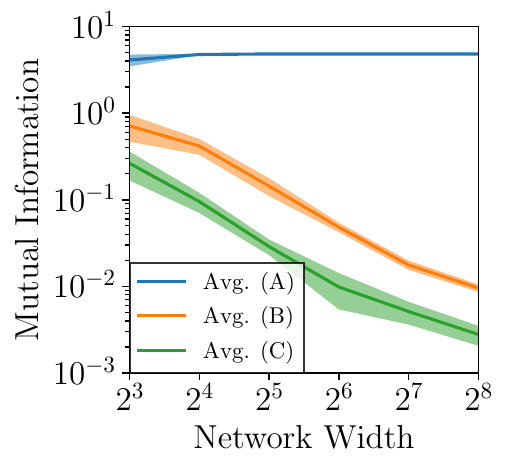}
	\caption{Result. }
	\label{fig:mutual_information_result}
	\end{subfigure}%
	\begin{subfigure}[t]{.3\columnwidth}
	\includegraphics[height=.9\columnwidth]{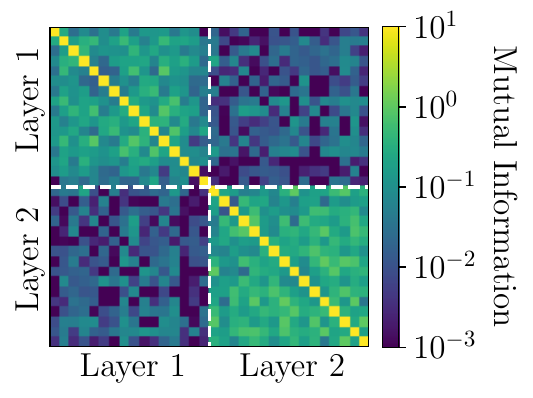}
	\caption{\textcolor{black}{Example.}}
	\label{fig:mutual_information_example}
	\end{subfigure}
	\caption{Nonlinear activations get statistically independent as the network width increases, supporting our assumption in Eq. \ref{eq:indep_activations}. Imagine a matrix containing mutual information between all pairs of nonlinear activations $h_k^l$ in a two-hidden-layer neural net. Group its entries into blocks as shown in panel (a): the diagonal (A) giving the entropy of an activation, the within-layer off-diagonal block (B) giving the dependence of sibling activations, the cross-layer off-diagonal (C) giving the dependence of activations in different layers. As seen in panel (b), the average mutual information in blocks (B) and (C) decreases sharply with increasing layer width.  Solid lines and shaded area represent average mutual information and its standard deviation over 100 repetitions. We show an example of the matrix with all pairwise mutual information values for a hidden layer width of 16 neurons in panel (c).
	}\label{fig:mutual_information}
\end{figure}

We test this assumption empirically by feeding a random input $x \sim N(0, I)$  into a neural net with two fully-connected and equally wide hidden layers with Dropout in between. We provide a detailed discussion on the Dropout layer in the context of NSDEs in Sec. \ref{subsec:dropout}.
Using the non-parametric entropy estimation toolbox, which is available \href{https://github.com/gregversteeg/NPEET}{here}\cite{npeet}, we can efficiently estimate the mutual 
information between all pairs of hidden layer activations. 
\textcolor{black}{
As depicted in Fig. \ref{fig:mutual_information_result} the mutual information between activations at different layers as well as within a layer shrinks fast when the hidden layers become wider. For a hidden layer width of 16 the mutual information between different layers is by a factor of $\sim$100 smaller than the average entropy in a layer. For a hidden layer width of practical use, such as 64 neurons, the mutual information between different layers is already by a factor of $\sim$1000 smaller than the average entropy in a layer. }  

Now the problem reduces to taking the expectations of the individual gradient terms. Despite being intractable, these expectations can be efficiently approximated by reusing the outcomes of the VMM step in Eq. \ref{eq:vmm} as follows
\begin{align}
\mathbb{E} [\nabla_{h_k^l} u_{l+1}(h_k^l)] &\approx \mathbb{E}  [\nabla_{\widetilde{h}_k^l} u_{l+1}(\widetilde{h}_k^l)] \label{eq:cross-cov-appx}\\
&=\int \nabla_{\widetilde{h}_k^l} u_{l+1}(\widetilde{h}_k^l) \mathcal{N}(\widetilde{h}_k^l|a_k^l, B_k^l) d \widetilde{h}_k^l.\nonumber
\end{align}

We attain a deterministic approximation to Stein's lemma  by taking the covariance $\Sigma_k$ from VMM and the expected gradient from Eq. \ref{eq:cross-cov-appx}:
\begin{align}
    &\mathrm{Cov}[x_k, f_\theta(x_k)] \!
    \approx\! \Sigma_k\! \prod_{l=0}^{L-1}\! \mathbb{E}_{\widetilde{h}_k^l}  [\nabla_{\widetilde{h}_k^l} u_{l+1}(\widetilde{h}_k^l)]\! := \!C_k. \label{eq:tractable_stein} 
\end{align}
We refer to applying this approximation throughout all neural net layers  as {\it Backward Vertical Moment Matching } (BVMM).
We provide the expected gradient $ \mathbb{E}_{\widetilde{h}_k^l} [\nabla_{\widetilde{h}_k^l} u_{l+1}(\widetilde{h}_k^l)]$ for commonly used layers in Sec. \ref{sec:layers}.

\subsection{The Bidimensional Moment Matching Algorithm}
\label{subsec:bmm}

Given an observed initial value  $x(t_n)$ at time point $t_n$, our deterministic method approximates the transition kernel $p(x(t_{n+1})|x(t_n))$ for an arbitrary $t_{n+1}$ by firstly discretizing the interval $t_{n+1}-t_n$ into $K$ steps such that $p(x(t_{n+1})|x(t_n)) \approx p(x_{K}|x_0)$.
Afterwards our method approximates $p(x_k|x_0)$ for  $k \in \{1, \ldots, K \}$ as below:
\begin{align}
\label{eq:bmm_transition_kernel}
       p(x_k|x_0) &=\int  p(x_{k}|x_{k-1})p(x_{k-1}|x_{0})   d x_{k-1}\\
        &=\int   \mathcal{N}  (x_{k}  |  m_{k}(x_{k-1}), S_{k}(x_{k-1}) )\nonumber\\
        &~~~~~~~~~\times  p(x_{k-1}|x_{0})   d x_{k-1}\nonumber\\
        &\approx\int   \mathcal{N}  (x_{k}  |  m_{k}(x_{k-1}), S_{k}(x_{k-1}) )\nonumber\\
        &~~~~~~~~~ \times \mathcal{N}(x_{k-1}|\mu_{k-1},\Sigma_{k-1})   d x_{k-1}\nonumber\\
        &\approx \mathcal{N}  (x_{k}  |  \mu_{k},\Sigma_k )\nonumber.
\end{align}
In the above expression, the moments of the normal density $\mathcal{N}(x_{k}|\mu_{k},\Sigma_{k})  $ at each time step $k$ are recursively calculated via HMM by applying the below moment matching rules in time direction:
\begin{align}
  \mu_{k+1} &:= \mu_k +  a_k^L \Delta t, \label{eq:final_next_mean}\\
  \Sigma_{k+1} &:= \Sigma_k + B_k^L  \Delta t^2 +(C_k + C_k^T) \Delta t \nonumber + D_k \Delta t,  
\end{align}
where $a_k^L \approx \mathbb{E} [ f_\theta(x_k) ] $ and $B_k^L \approx \mathrm{Cov}[ f_\theta(x_k) ]$  \textcolor{black}{are approximated via VMM as defined in Sec. \ref{subsec:drift_moments}. 
The term $C_k \approx \mathrm{Cov}[x_k, f_\theta(x_k)]  $ is obtained \textcolor{black}{ via Eq. \ref{eq:tractable_stein}}. 
The second central moment of the diffusion function $D_k \approx \mathbb{E}[{L}_{\phi}(x_k){L}_{\phi}^T({x}_k)]$ is approximated via Eq. \ref{eq:diff_central_moment}}. 
We refer to our method as {\it Bidimensional Moment Matching }(BMM) and provide its pseudocode in Algorithm \ref{alg:main}.

\begin{algorithm}
    \scriptsize
	\caption{Bidimensional Moment Matching (BMM)}\label{alg:main}
	\begin{algorithmic}
	\State {\bf Inputs:}  ${f}_{{\theta}}(\cdot) :=  u_L ( u_{L-1} ( \ldots u_2( u_1(\cdot) ) \ldots ) )$ \Comment{Drift Net}
	\State ~~~~~~~~~~~~~  ${L}_{{\phi}}(\cdot) :=  v_L ( v_{L-1} ( \ldots v_2( v_1(\cdot) ) \ldots ) )$  \Comment{Diffusion Net}
	\State ~~~~~~~~~~~~~ $x(t_n)$ \Comment{Initial Value}
	\State ~~~~~~~~~~~~~ $t_{n+1}-t_n$ \Comment{Time Horizon}
	\State ~~~~~~~~~~~~~ $K$ \Comment{Discretization Steps}
	\State {\bf Outputs:} Approximate transition kernel $p(x_K|x_0)$
	\State $\Delta t \leftarrow (t_{n+1}-t_n)/K$ \Comment{Discretize}
	\State $\mu_0, \Sigma_0 \leftarrow x_0, I\epsilon$ \Comment{Initialize, with $\epsilon \in \mathbb{R}_+$ being a small number}
       \For{ time step $k \in \{0,\cdots,K-1\}$ } \Comment{Horizontal Moment Matching}
       \State $\widetilde{h}_k^0, \widetilde{e}_k^0 \leftarrow \mathcal{N}(\mu_k, \Sigma_k)$ \Comment{Input distribution to drift and diffusion net}
       \For{layer index $l \in \{1,\cdots,L\}$ } \Comment{Vertical Moment Matching}
       \State \textit{See Sec. \ref{sec:layers} for expectation, covariance, and Jacobian of $u_l$ and $v_l$  }
       \State $a_k^l \leftarrow \mathbb{E}[u_l(\widetilde{h}_k^{l-1})]$ \Comment{Drift net, expectation at layer $l$}
       \State $B_k^l \leftarrow \mathrm{Cov}[u_l(\widetilde{h}_k^{l-1})]$ \Comment{Drift net, covariance at layer $l$}
       \State $\widetilde{h}_k^l \leftarrow \mathcal{N}(a_k^l, B_k^l)$  \Comment{Drift net, output distribution at layer $l$}
       \State $J_k^l \leftarrow \mathbb{E} [\nabla_{\widetilde{h}_k^{l-1}} u_l(\widetilde{h}_k^{l-1})]$      \Comment{Drift net, Jacobian at layer $l$}
       \State $c_k^l \leftarrow  \mathbb{E}[v_l(\widetilde{e}_k^{l-1})]$   \Comment{Diffusion net, expectation at layer $l$}
       \State $D_k^l \leftarrow \mathrm{Cov}[v_l(\widetilde{e}_k^{l-1}) ]$  \Comment{Diffusion net, covariance at layer $l$}
       \State $\widetilde{e}_k^l \leftarrow \mathcal{N}(c_k^l, D_k^l)$  \Comment{Diffusion net, output distribution at layer $l$}
       \EndFor
	\State $D_k \leftarrow (D_k^L + (c_k^L)(c_k^L)^T) \odot I $ \Comment{Diffusion net,  second central moment}
	   \State $C_k \leftarrow  \Sigma_k \prod_{l=1}^L J_k^l$ \Comment{Stein's lemma}
        \State $\mu_{k+1} \leftarrow \mu_k +  a_k^L \Delta
        t$\Comment{Mean and covariance at time step $k+1$}
        \State $\Sigma_{k+1} \leftarrow\Sigma_k + B_k^L  \Delta t^2 +(C_k + C_k^T) \Delta t \nonumber + D_k \Delta t $ 
        \State $p({x}_{k+1}|x_0) \leftarrow \mathcal{N}({x}_{k+1}|\mu_{k+1},\!  \Sigma_{k+1})$  \Comment{Transition kernel at  $k+1$}
        \EndFor
        \State \Return $p(x_K|x_0)$
\end{algorithmic}
\end{algorithm}

%% file: sections/layers.tex
\section{Moments of layers and their derivatives}
\label{sec:layers}
Given the VMM output of the previous layer as $\widetilde{h}_k^l \sim \mathcal{N}(a_k^l, B_k^l)$, we show below how the output moments and expected Jacobian can be calculated for three common  layer types: (i) linear activations, (ii) nonlinear activations, and (iii) Dropout.
Additionally we provide a discussion about the implications of the Dropout layer and how it can be used to make BMM a tighter approximation.

\subsection{Linear Activations} 
The  moments of an affine transformation $u_{l+1}( \widetilde{h}_k^{l}) =  W^{l+1}  \widetilde{h}_k^l +  b^{l+1}$  are analytically available as 
\begin{align}
    \mathbb{E}[u_{l+1}( \widetilde{h}_k^{l})]&=  W^{l+1}  a_k^{l} +  b^{l+1},\\ 
    \mathrm{Cov}[u_{l+1}( \widetilde{h}_k^{l})]&=  W^{l+1}   B_k^{l}  (W^{l+1})^T, \nonumber
\end{align}
where $W^{l+1}$ and $b^{l+1}$ correspond to the weights and bias of layer $l+1$. The expected Jacobian is a constant
\begin{equation}
    \mathbb{E}\left[\nabla _{\widetilde{h}_k^l } u_{l+1}(\widetilde{h}_k^{l})    \right] =  W^{l+1}.
\end{equation}

\subsection{Nonlinear Activations} 
The output moments of nonlinear activations are analytically not  tractable. However, for many types of nonlinearities in widespread use exist tight approximations. 
For instance, the ReLU moments can be approximated as \cite{wu2018deterministic}
\begin{align}
     \mathbb{E}[\max (0,  \widetilde{h}_k^l)]\approx &\sqrt{\text{diag}( B_k^l)}  {SR}\left( a_k^l/\sqrt{\text{diag}( B_k^l)}\right), \nonumber\\
     \mathrm{Cov}[\max (0,  \widetilde{h}_k^l)] \approx &\sqrt{\text{diag}( B_k^l) ( B_k^l)^T}{F}( a_k^{l},  B_k^{l} ),
\end{align}
where $\text{SR}(x)= \phi(x) +  x \Phi(x)$ with
 $\phi$ and $\Phi$ representing the PDF and CDF of a standard Gaussian variable and 
 \begin{equation}
{F}( a^{l}_k,  B^{l}_k )=\left( A( a^{l}_k,  B^{l}_k )+ \exp{- Q(a^{l}_k,  B^{l}_k)} \right).
\end{equation}
In order to keep the paper self contained we detail the functions $A$ and $Q$ below and refer to \cite{wu2018deterministic} for the derivation.
After introducing the dimensionless vector $\epsilon^l_k= a^l_k/\sqrt{\text{diag}( B^l_k)}$,
the function $A( a^{l}_k,  B^{l}_k )$ is estimated as
\begin{equation}
     A( a^{l}_k,  B^{l}_k ) = \text{SR}( \epsilon^l_k)\text{SR}(\epsilon^l_k)^T +  \rho^l_k \Phi( \epsilon^l_k)\Phi(\epsilon^l_k)^T,
\end{equation}
with $ \rho^l_k = B^l_k/ \left(\sqrt{\text{diag}(B^l_k)}\sqrt{\text{diag}(B^l_k)^T} \right)$.
The $i,j$-th element of $ Q(a^{l}_k,  B^{l}_k)$ can be estimated as:
\begin{align}
     Q(a^{l}_k,  B^{l}_k)_{i,j} =  
    &\frac{\rho^l_{k_{i,j}}}{2 g^l_{k_{i,j}} (1+\bar{\rho}_{k_{i,j}})} \left( (\epsilon_{k_i}^l)^2 +(\epsilon_{k_j}^l)^2 \right)-\\
    &\frac{\arcsin(\rho^l_{k_{i,j}}) - \rho^l_{k_{i,j}}}{\rho_{k_{i,j}} g^l_{k_{i,j}}}\epsilon_{k_i}^l\epsilon_{k_j}^l-\log \left( \frac{g^l_{k_{i,j}}}{2 \pi} \right), \nonumber
\end{align}
with $ g^l_k=\arcsin( \rho^l_k) +  \rho^l_k \oslash (1+ \bar{ \rho}^l_k)$, and
$\bar{\rho}^l_k = \sqrt{1- \rho^l_k \odot  \rho^l_k}$. We denote with $\oslash, \odot$ elementwise division, multiplication.

Since activation functions are applied element-wise, off-diagonal entries of the expected gradient are zero.  The diagonal of the Jacobian of the ReLU function is the expected  Heaviside step function \cite{wu2018deterministic}
\begin{align}
  \text{diag}\left(\mathbb{E}\left[\nabla _{\widetilde{h}_k^l } u_{l+1}(\widetilde{h}_k^{l})    \right]\right) \approx
    \Phi\left( a_k^l/\sqrt{\text{diag}( B_k^l)}\right).
\end{align}

\subsection{Dropout} 
\label{subsec:dropout}
Dropout is defined as the mapping $u_{l+1}(\widetilde{h}_k^l) := \widetilde{h}_k^l \odot \beta_k^l/q$ for a random vector $\beta_k^l$ consisting of $\text{Bernoulli}(q)$ distributed entries. 
The moments of which are available as
\begin{align}
     \mathbb{E}[u_{l+1}(\widetilde{h}_k^l)] &=  a_k^l,  \\
     \mathrm{Cov}[u_{l+1}(\widetilde{h}_k^l)] &=  B_k^{l}+ \text{diag}\left( \frac{1-q}{q}\left( B_k^l + ( a_k^l) ( a_k^l)^T \right)  \right). \nonumber
\end{align}

We obtain the first moment by using the independence between $ \widetilde{h}^l_k$ and $  \beta^l_k$ 
\begin{align*}
    \mathbb{E}[  u_{l+1}(\widetilde{h}^l_k)]   = \mathbb{E}[  \widetilde{h}^l_k \odot \beta^l_k /q] 
                              = \mathbb{E}[   \widetilde{h}^l_k]\mathbb{E}[  \beta^l_k]/q
                              = \mathbb{E}[h^{l}_k].   
\end{align*}
We derive  diagonal and off-diagonal entries in $  \mathrm{Cov}[u_{l+1}(\widetilde{h}^l_k)]$ separately. We obtain for $i=j$
\begin{align}
    \mathrm{Cov}[u_{l+1}(\widetilde{h}^l_k)]_{i,i}&
    =\mathbb{E}\left[(u_{l+1}(\widetilde{h}^l_k)_i)^2 \right] - 
                              \mathbb{E}[u_{l+1}(\widetilde{h}^l_k)_i]^2\\\nonumber
                             &=\frac{1}{q^2}\mathbb{E}\left[(\widetilde{h}^l_{k_i})^2 \right]\mathbb{E}\left[(\beta^l_{k_i})^2 \right] -\mathbb{E}[\widetilde{h}^l_{k_i}]^2\\\nonumber
                             &= \mathbb{E}\left[(\widetilde{h}^l_{k_i})^2 \right] -\mathbb{E}[\widetilde{h}^l_{k_i}]^2 +\nonumber \frac{1-q}{q}\mathbb{E}\left[(\widetilde{h}^l_{k_i})^2 \right] \\
                             &=  (B_k^{l})_{i,i}+ \frac{1-q}{q}\left( B_k^l + ( a_k^l) ( a_k^l)^T \right)_{i,i}\nonumber
\end{align}
and for $i\neq j$
\begin{align}
     \mathrm{Cov}[u_{l+1}(\widetilde{h}^l_k)]_{i,j}
     &=\mathbb{E}\left[u_{l+1}(\widetilde{h}^l_k)_i u_{l+1}(\widetilde{h}^l_k)_j \right] -\\ \nonumber
                              &~~~~~\mathbb{E}[u_{l+1}(\widetilde{h}^l_k)_i]\mathbb{E}[u_{l+1}(\widetilde{h}^l_k)_j]\\\nonumber
                              &=\frac{1}{q^2} \mathbb{E}\left[\widetilde{h}^l_{k_i} \widetilde{h}^l_{k_j} \beta_{k_i}^l \beta_{k_j}^l\right]-\mathbb{E}[\widetilde{h}^l_{k_i}]\mathbb{E}[\widetilde{h}^l_{k_j}]\\\nonumber
                              &= \mathbb{E}\left[\widetilde{h}^{l}_{k_i} \widetilde{h}^{l}_{k_j}\right]  -\mathbb{E}[\widetilde{h}^l_{k_i}]\mathbb{E}[\widetilde{h}^l_{k_j}]\\\nonumber
                              &=   (B_k^{l})_{i,j}.
\end{align}
The Jacobian of the Dropout layer is  the identity matrix\begin{equation}
\mathbb{E}\left[\nabla _{\widetilde{h}_k^l } u_{l+1}(\widetilde{h}_k^{l})    \right] =\mathbb{E}\left[ I \beta_k^l / q   \right] = I.
\end{equation}

\subsubsection{Dropout tightens the Gaussian assumption}
\begin{figure}[h]
	\centering
	\begin{subfigure}[t]{.55\columnwidth}
	\centering
		\includegraphics[height=.43\columnwidth]{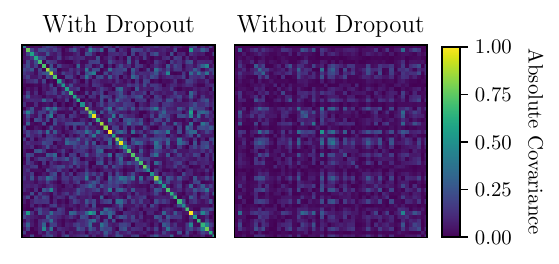}
		\caption{\textcolor{black}{Intermediate Activation.}}
		\label{fig:dropout_activation}
	\end{subfigure}%
	\begin{subfigure}[t]{.45\columnwidth}
	\includegraphics[height=.53\columnwidth]{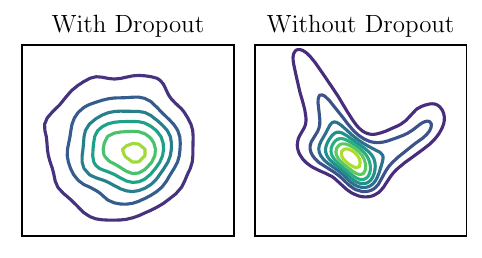}
	\caption{Output Distribution. }
	\label{fig:dropout_output}
	\end{subfigure}
	\caption{Dropout \textcolor{black}{reduces the correlation coefficient between different activations}. We pass a multivariate normal distributed random vector through a neural net with three 50-neuron-wide hidden layers with ReLU activation. The off-diagonals of the covariance matrix of the activation map are suppressed when Dropout is used after each ReLU activation, as shown in Panel (a) for an intermediate layer and Panel (b) for the output layer. Decorrelation of a large number of co-variates makes a normal distribution an accurate approximation on their sum due to the central limit theorem.
	}\label{fig:dropout}
\end{figure}

\textcolor{black}{
As shown in the above equations, Dropout increases the value of the diagonal entries in the covariance matrix relative to its off-diagonal entries. 
Consequently, Dropout is helpful to reduce the correlation coefficient between different activations in the same layer, making their sum approach the normal distribution due to the \textit{Central Limit Theorem (CLT)}.
As Dropout is added, we see in Fig. \ref{fig:dropout_activation} the covariance matrix at a hidden layer to be dominated by its diagonal values, which results in an approximately Gaussian output as shown in Fig. \ref{fig:dropout_output}. 
Though Dropout does not strictly guarantee total decorrelation, i.e. negligible  off-diagonal covariances compared to diagonal ones, we observe in our experiments this assumption to be valid for 
\textcolor{black}{neural networks layer widths of practical relevance.}
As discussed in Sec. \ref{subsec:cross_cov}, we observe for an increasing layer width decreasing mutual information between activations in the same layer as well as between different layers. 
We visualize in Fig. \ref{fig:mutual_information_result} the mutual information between different neurons for varying layer widths. For a layer width of 16, the mutual information within a layer is $\sim$10 times smaller than the average entropy in the layer and is  $\sim$100 times smaller for \textcolor{black}{neural nets with layers having more than 64 neurons.} As shown in Fig. \ref{fig:mutual_information_example} the mutual information is dominated by its diagonal values for a layer width of 16.}

\subsubsection{Dropout as an augmented NSDE}
Stein's lemma assumes a deterministic drift function, which firstly contradicts the usage of Dropout layers. 
By reinterpreting the Dropout units as augmented states \cite{aug_ode} we obtain a valid form, i.e. a deterministic drift function.

We collect the Dropout units from all layers in the variable $\Tilde{\beta}(t) = [\beta^1(t), \ldots, \beta^L(t)]$ and define the augmented state variable $\Tilde{x}(t)$ as
\begin{equation}
    \Tilde{x}(t) =
    \begin{bmatrix}
   x(t)\\ \Tilde{\beta}(t)
   \end{bmatrix} 
\end{equation}
with augmented dynamics
\begin{equation}
    d \Tilde{x}\!=\!\begin{bmatrix}
       f_{\theta}(x(t), r(\Tilde{\beta}(t)) ) d t  +L_{\phi}(x(t), r(\Tilde{\beta}(t))) d w_x(t)\\
       -0.5 \Tilde{\beta}(t) dt + d w_{\Tilde{\beta}}(t)
    \end{bmatrix}.
\end{equation}
Above $x(t)$ follows an NSDE, as described in Eq. \ref{eq:nsde}, with additional input $\Tilde{\beta}(t)$ and modification function $r$. The Dropout units follow an Ornstein-Uhlenbeck process with stationary distribution. \textcolor{black}{If the stationary distribution exists, a NSDE converges to it as time goes to infinity.
Once the stationary distribution is reached, it will not change any further as time progresses. 
Here, the stationary distribution of the Dropout units exists and takes the form
$\lim_{t \rightarrow \infty} p(\Tilde{\beta}(t))= \mathcal{N}(0, I)$ \cite{applied_sde}.
For Dropout units, the distribution does not change as time progresses, and we can assume that the distribution at each time point follows their stationary distribution starting form the initial time point $t_0$.
}
The modification function $r$ maps $\Tilde{\beta}(t)$, which is distributed accordingly to a standard Gaussian, to Bernoulli distributed random variables as 
\begin{equation}
    r(\Tilde{\beta}(t)) = \text{sgn}(\Phi(\Tilde{\beta}(t))-(1-q)),
\end{equation}
where $\Phi$ is the CDF of a standard Gaussian, $\text{sgn}$ is the signum function, and $q$ is the parameter of the Bernoulli distribution, i.e. the probability of drawing a $1$. 
The modification function $r$ and the dynamics of the Dropout units contain no learnable parameters.
Given this reinterpretation of the Dropout layer, drift and diffusion networks are deterministic functions of their inputs.

%% file: sections/numerical_experiments.tex
\section{Numerical Properties}
\label{sec:numerical_properties}

We investigate the numerical properties of the BMM algorithm in terms of integration error, approximation error of the expected Jacobian, computation cost, and generalization to multiple modes. 
We  compare cubature as an alternative choice to VMM, which is a standard numerical method for approximating the expectation of a smooth function with respect to a normal distribution.  
Our empirical findings demonstrate that VMM has favorable computational properties over cubature.

\subsection{Cubature}
Cubature estimates the expected value of a nonlinear function $f_\theta(x)$  with respect to a Gaussian density  $\mathcal{N}(x|\mu, \Sigma)$ as  a weighted sum of point mass evaluations 
\begin{align}
\int f_\theta(x)\mathcal{N}({x}|\mu, \Sigma)dx \approx
\sum_{i=1}^U w_i {f_\theta}(\mu + \sqrt{\Sigma}{\zeta}_i),
\label{eq:qmc}
\end{align}
where  \textcolor{black}{$\sqrt{\Sigma}$ is the square root of the covariance matrix and routineley computed by using the Cholesky decomposition.} The coefficients $w_i$ and ${\zeta}_i$ are predetermined by a heuristic that aims to spread the particles in a maximally information preserving way. 
There exist multiple heuristics for choosing $w_i$ and ${\zeta}_i$.  
In this paper we use the commonplace heuristic \textit{Unscented Transform} (UT) \cite{Wan00theunscented}, which evaluates the above expression  as a sum of $U=2D+1$ elements for a $D-$dimensional input space.

 \begin{figure*}[h]
	\centering
	\begin{subfigure}[t]{.75\columnwidth}
	\includegraphics[height=3cm]{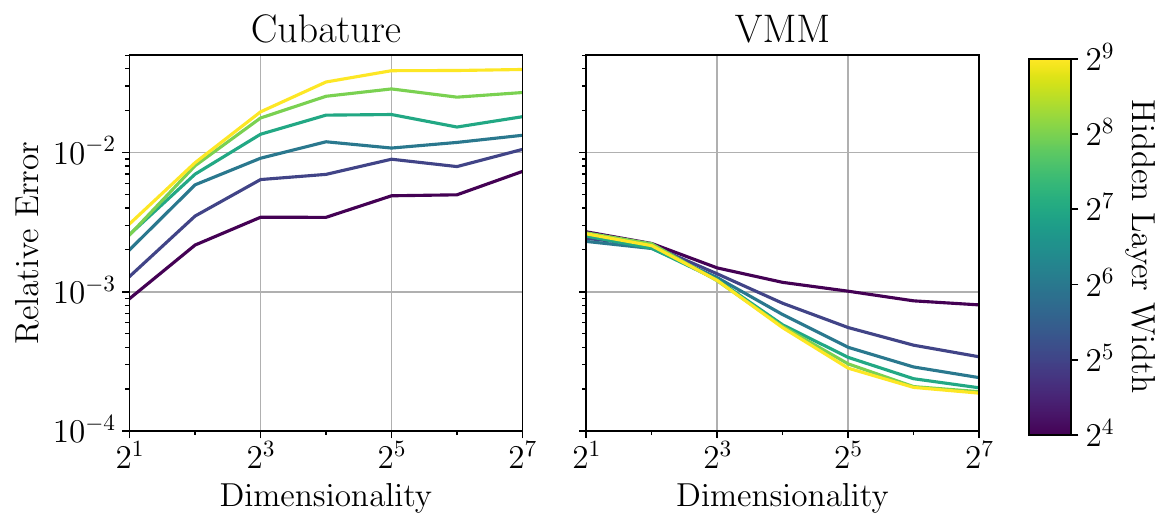}
		\caption{Integration error comparison.}
		\label{fig:integration}
	\end{subfigure}
	 \hfill
	\begin{subfigure}[t]{.75\columnwidth}
	\includegraphics[height=3cm]{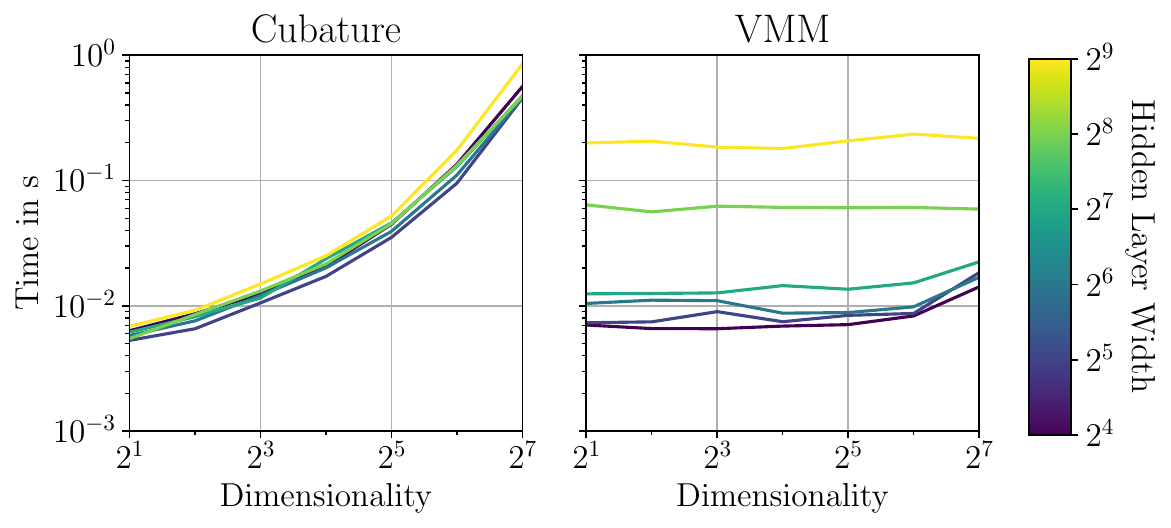}
	\caption{Wall clock time comparison.}
	\label{fig:timing}
	\end{subfigure}
	\hfill
	\begin{subfigure}[t]{.5\columnwidth}
	\centering
	\includegraphics[height=3cm]{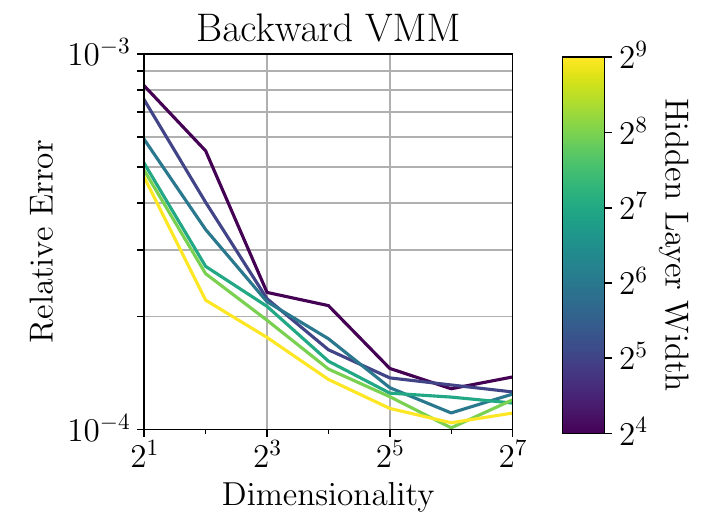}
	\caption{Approximation error of the expected Jacobian.}
	\label{fig:jacobian}
	\end{subfigure}
	\caption{Comparison of VMM versus cubature in terms of relative error and computation time. We generate a randomly initialized neural net $f_{\theta}(x)$. The dimensionalities of $x$ and $f_{\theta}(x)$ are equal and vary in the horizontal axis of all plots. The neural net $f_{\theta}(x)$ has three fully-connected layers of varying widths color-coded according to the heatmap on the right of the figures, ReLU activation, and Dropout with rate 0.2. As cubature cannot handle a random $f_\theta(x)$ and discontinuous activations, we evaluate it with tanh activation and without Dropout. We aim to approximate the intractable expectation $r=\int f_{\theta}(x) \mathcal{N}(x|{\mu}, {\Sigma}) dx$, where ${\mu} \sim \mathcal{N}(0, I)$ and  ${\Sigma} \sim \mathcal{W}( I, \mathrm{dim}( I))$ with Wishart distribution $\mathcal{W}$. We repeat the experiment 512 times and report the average relative error $||r-\hat{r}||^2/||r||^2$ in Panel (a), where $r$ is represented by averaging over $10$ million Monte Carlo simulations and $\hat{r}$ is approximated via cubature and VMM, respectively. We calculate the computation time of this experiment in all repetitions and report its average as a function of input dimensionality in Panel (b).
	In Panel (c) we report the average relative error of the true Jacobian $\mathbb{E}[\nabla_x f_\theta(x)]$, which is obtained by the Monte Carlo simulation, and our approximate Jacobian obtained by backward   vertical   moment   matching.}
\label{fig:numerical_tests}
\end{figure*}


\subsection{Approximation Error.}
In Fig. \ref{fig:integration}, we compare VMM and cubature in approximating $\mathbb{E} [ f_\theta(x) ]$ for a normal distributed input $x$ and the effect of neural net width and input/output dimensionality on approximation accuracy for a randomly initialized neural net $f_{\theta}(x)$. For low dimensionalities, the relative error of VMM is approximately equal to cubature. The approximation error of VMM shrinks with increasing hidden layer width and dimensionality. This is expected since summing a larger number of decorrelated variables makes the assumption of normally distributed intermediate activations 
more accurate due to the central limit theorem \cite{fastdropout}. 
Similarly, we observe the approximation error of the expected Jacobian by Backward VMM to decrease with increasing input dimensionality and hidden layer width as shown in Fig. \ref{fig:jacobian}.


\subsection{Computational Cost.} 
\label{subsec:computational_cost}
\textcolor{black}{
Propagation of $S$ particles with dimensionality $D$ along $K$ time steps requires $\mathcal{O}(SKHD + SKH^2)$ computations, when dynamics is governed by a NSDE with hidden layer width $H$. 
The first term  is due to the computational cost of the $H\times D$-dimensional affine transformation in the first layer. The second term $\mathcal{O}(SKH^2)$ is due to the computational cost of the $H\times H$-dimensional affine transformations in the subsequent hidden layers, which require $\mathcal{O}(H^2)$ computations.
Our method BMM approximates the $S \rightarrow \infty$ limit, while requiring only $\mathcal{O}(KHD^2 + KH^2D + KH^3)$ computations. 
The first two terms are due to the computational cost of the $H\times D$-dimensional affine transformation in the first layer. The third term is due to the computational cost of the $H\times H$-dimensional affine transformations in the subsequent hidden layers.
Replacing VMM with cubature in our framework results in an algorithm requiring  $\mathcal{O}(KHD^2+KH^2D+KD^3)$ computations. 
Cubature requires at least $\mathcal{O}(D)$ NSDE evaluations, which causes the additional factor $D$ in the first two terms. The third term arises from the Cholesky decomposition of the input, as shown in Eq. \ref{eq:qmc} during point selection.}
We visualize in Fig. \ref{fig:timing} the wall clock time of VMM and cubature as a function of dimensionality and hidden layer width.
Dimensionality sets a bottleneck for cubature, while it barely affects VMM. 
Contrarily, VMM gets significantly slower as the hidden layer width increases, while the computational cost of cubature remains similar. VMM is adaptable to setups requiring high learning capacity by building narrow and deep architectures. However, dimensionality is an external factor that limits the applicability of cubature.
\textcolor{black}{As shown in Fig. \ref{fig:timing}, VMM has a runtime of approximately 10\,ms for an input dimensionality and hidden layer width up to 128 for architectures with three hidden layers, which is sufficient for most applications. In case that larger neural network architectures are required, the runtime can be reduced by sparse covariance approximations. We leave the investigation of sparse covariance approximations as a promising direction of future work.}

\subsection{Multimodal Processes.} 
\begin{figure}[h]
	\centering
	\begin{subfigure}[t]{0.5\columnwidth}
		\centering
		\includegraphics[width=1\columnwidth]{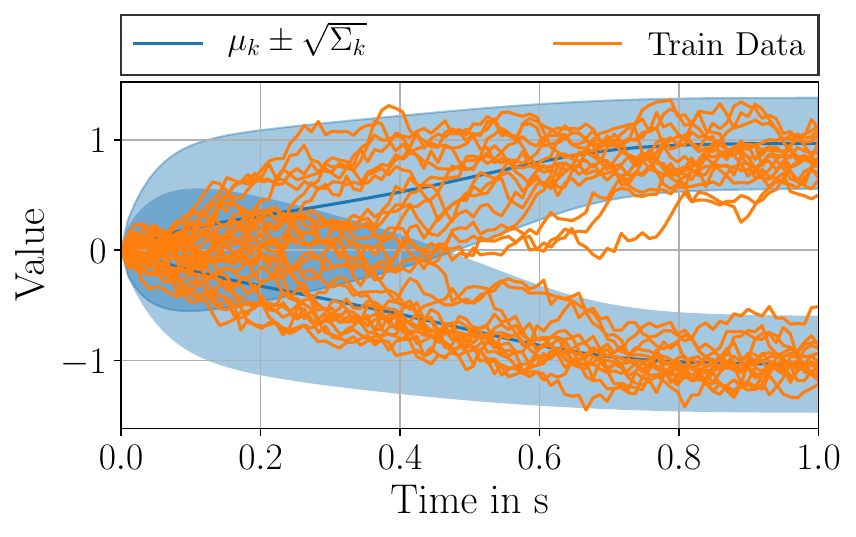}
		\caption{Predictions}
		\label{fig:dw_dinsde}
	\end{subfigure}%
    \hfill
	\begin{subfigure}[t]{0.5\columnwidth}
	\centering
	\includegraphics[width=1\columnwidth]{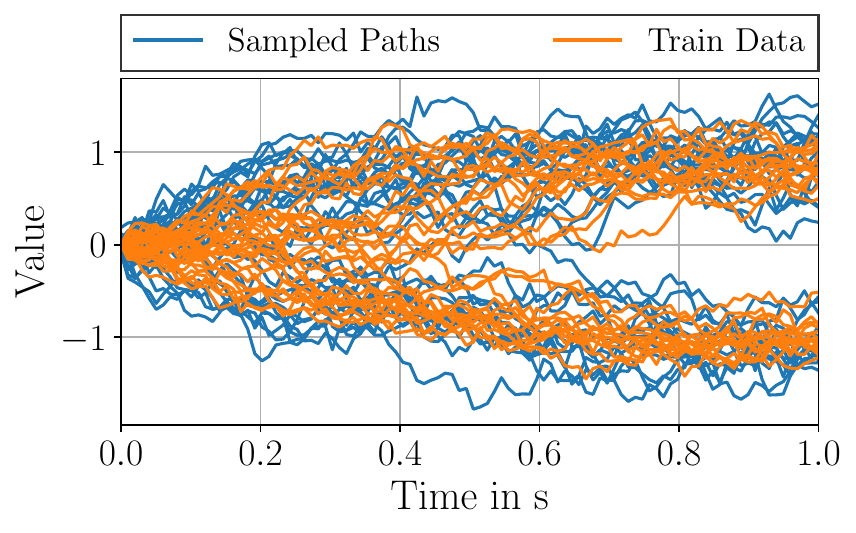}
	\caption{Sampled paths}
	\label{fig:dw_em}
	\end{subfigure}
	\caption{Handling multimodality with a NSDE applied to double-well potential dynamics: $d x_t = 4 x_t (1- x_t^2) dt + dw_t$. We train a separate model on each mode and predict with BMM in Panel (a), and with NSDE-MC in Panel (b).
	}\label{fig:dw_all}
\end{figure}

Deterministic prediction of unimodal densities with NSDEs is an unsolved problem of its own. We restrict our focus on unimodal solutions, as the first inevitable step towards multimodal solutions. That being said, our method can generalize to multiple modes under mild assumptions. For instance, if training sequences come with the ground-truth knowledge of the modality they belong to, a separate unimodal NSDE can be fit to each mode and their mixture can be used during prediction. If modality assignments are not known a-priori, an initial clustering step can be applied. We visualize  prediction results in Fig. \ref{fig:dw_all} on the bimodal double-well dynamics, after clustering the training data and training a separate NSDE on each mode. In cases calling for more advanced solutions, BMM can serve as a subroutine in a Bayesian model, such as deterministic variational inference of a Dirichlet process mixture model \cite{dpmog} or 
by introducing an auxiliary latent variable, which determines the mode \cite{multiple_futures}.

%% file: sections/experiments.tex
\section{Benchmarks}
\label{sec:experiments}
\textcolor{black}{
We demonstrate in \textcolor{black}{four} different applications that NSDEs can be used for a broad range of scenarios}.
In the first experimental section, we use NSDEs as a continuous depth layer for neural networks.
We demonstrate the performance of NSDEs as a continuous depth layer  on eight different UCI datasets.
\textcolor{black}{In the second experimental section, we benchmark NSDEs on time series classification tasks on the MNIST and IMDB dataset}.
In the third experimental section, we use NSDEs for time series modeling. 
We benchmark on three different time series prediction datasets that enable a comparison to the state of the art methods for learning-based modeling of dynamics.
\textcolor{black}{In the fourth experimental section, we compare our method against Monte Carlo sampling for modeling high dimensional dynamics.
}

\subsection{Evaluation Criteria} 
The prediction quality is assessed in terms of \textit{mean squared error} (MSE),  \textit{negative log-likelihood } (NLL), \textit{expectation of coverage probability error} (ECPE),
\textcolor{black}{ as well as the  \textit{expected calibration error } (ECE). Both ECPE and ECE measure the calibration of the uncertainty estimates of our model, i.e. how well the predictive distribution covers the true data distribution. The ECPE is a suitable metric when the target variable is continuous, while the ECE is suitable when the target is discrete, i.e. classification tasks.}
ECPE  measures the absolute difference between true confidence and the empirical coverage probability as \cite{calibrated_regression}
\begin{equation}
    ECPE = \frac{1}{J} \sum_{j=1}^{J} |\hat{p}_j - p_j|,
    \label{eq:ecpe}
\end{equation}
where $p_j$ and $\hat{p}_j$ is the true frequency and  empirical frequency, respectively. 
Loosely speaking, ECPE is small if $p$ percent of the data lies in the predicted $p$-percent confidence interval.
We choose 10 equally spaced confidence levels between 0 and 1.
By taking the average over all test samples $x_i \in \mathcal{D}_{test}$, which lie in a predicted confidence interval, the empirical frequency is estimated as 
\begin{equation}
    \hat{p}_j = \frac{\sum_i^{|\mathcal{D}_{test}|} \mathbb{I}\{ x_i \leq \hat{F}_i^{-1}(p_j)\}}{|\mathcal{D}_{test}|}.
    \label{eq:def_calibration}
\end{equation}
In contrast to \cite{calibrated_regression}, we consider in our work the case of multivariate predictors, which complicates the estimation of the predicted inverse cumulative distribution function  $\hat{F}_i^{-1}$.
However, if $x_i$ is normally distributed, we can analytically estimate $\hat{F}_i^{-1}$ as a function of model outputs  $\mu_i$ and $\Sigma_i$ 
As discussed by \cite{2013NDIMENSIONALCF}, we may define the cumulative distribution function as the probability that a sample lies inside the ellipsoid determined by its Mahalanobis distance.
The ellipsoidal region is analytically obtained as
\begin{equation}
  (x_i -  \mu_i )^T  \Sigma_i^{-1}( x_i -  \mu_i ) \leq \chi_{D}^2(p) = \hat{F}_i^{-1}(p),
\end{equation}
where $\chi_{D}^2$ is the chi-squared distribution with $D$ degrees of freedom.

\textcolor{black}{
The ECE is commonly used for assessing the calibration quality for classification tasks. 
It measures the difference between the fraction of predictions that are correct (accuracy) and the mean of the probabilities (confidence) for a probability interval, called a bin.  We use $J=10$ equally spaced bins. 
As an example, bin number six contains all predictions within the probability range $[0.5, 0.6)$.
We use the calculation procedure as proposed in \cite{ece}
\begin{equation}
    ECE = \sum_{j=1}^{J} \frac{|B_j|}{N}|acc(B_j) - conf(B_j)|, 
\end{equation}
where $N$ is the number of samples, $B_j$ is the set of predictions whose prediction confidence falls into the $j$-th bin, $acc(B_j)$ is the average accuracy in the $j$-th bin, and  $conf(B_j)$ is the average confidence in the $j$-th bin.}

\subsection{NSDEs as Continous Depth Layers}

\begin{table*}[!h]
	\caption{Negative log likelihood values of 8 benchmark datasets. We report average and standard error over 20 runs.}
	\label{tab:uci_res_nll}
	\centering
	\resizebox{0.9\textwidth}{!}{
		\begin{tabular}{ l c  cc cc cc cc}
		\toprule
			& & Boston & Energy & Concrete & Wine Red & Kin8nm & Power & Naval & Protein \\
			\cmidrule(lr){3-10}
			& $N$ & 506 & 768 & 1,030  & 1,599 & 8,192 & 9,568 & 11,934 & 45,730  \\
			& $D$ & 13 & 8 & 8 & 22 & 8 & 4 & 26& 9  \\
			\midrule
				
			Dropout \cite{dropout_gal} && 2.46(0.06) & 1.99(0.02) & {3.04(0.02)} & 0.93(0.01)& -0.95(0.01) &
			\textbf{2.80(0.01)}&-3.80(0.01)&2.89(0.00)\\
			DVI \cite{wu2018deterministic}&& {2.41(0.02)} & 1.01(0.06) & 3.06(0.01) & \textbf{0.90(0.01)} & {-1.13(0.00)} & \textbf{2.80(0.00)} & \textbf{-6.29(0.04)}& 2.85(0.00)\\
			 \midrule
			NSDE-MC \cite{adjoint_sde}\\
			Integration Time: 2s && 2.62(0.04) & 1.63(0.05) & 3.19(0.03) & 1.01(0.01)&-1.15(0.01) & 2.97(0.01)&-4.01(0.01)& 2.88(0.01)\\
			Integration Time: 4s && 2.60(0.05) & 1.33(0.08) & 3.22(0.03) & 1.01(0.01)&-1.12(0.01) & 2.99(0.01)&-4.01(0.02)& 2.89(0.01)\\
			\textcolor{black}{Integration Time: 8s} && \textcolor{black}{2.66(0.07)} & \textcolor{black}{0.96(0.06)} & \textcolor{black}{3.24(0.02)} & \textcolor{black}{1.06(0.01)} & \textcolor{black}{-1.06(0.01)} & \textcolor{black}{3.01(0.01)} & \textcolor{black}{-4.00(0.02)} & \textcolor{black}{2.93(0.01)} \\
			 \midrule
			NSDE-Cubature (Ours, Ablation)  \\
			Integration Time: 2s && 2.74(0.05) & 1.79(0.02)  & 3.31(0.02) &0.98(0.01)&-0.89(0.00) & 2.85(0.01)&-3.87(0.01)& 2.92(0.00)\\ 
			Integration Time: 4s && 2.63(0.05) & 1.45(0.02)  & 3.30(0.02) &0.98(0.01)&-0.98(0.01) & 2.85(0.01)&-4.18(0.04)& 2.89(0.01)\\
			\textcolor{black}{Integration Time: 8s} && \textcolor{black}{2.56(0.08)} & \textcolor{black}{1.42(0.02)}  & \textcolor{black}{3.28(0.02)} & \textcolor{black}{0.99(0.01)}& \textcolor{black}{-0.97(0.01)} & \textcolor{black}{2.86(0.01)} & \textcolor{black}{-4.32(0.07)} & \textcolor{black}{2.90(0.00)}\\
			  \midrule
			  NSDE-BMM (Ours, Proposed)\\
			  Integration Time: 2s && 2.45(0.02) & 1.18(0.07) & 3.00(0.01) & 0.97(0.01) &-1.14(0.00) & 2.81(0.01) & -3.92(0.01)& 2.85(0.00)\\
			  Integration Time: 4s && {2.41(0.02)} & {0.82(0.06)} &{2.92(0.02)}  & {0.96(0.01)} &-1.21(0.01) & {2.81(0.01)} & -4.01(0.01)& {2.77(0.01)}\\
			  \textcolor{black}{Integration Time: 8s} && \textcolor{black}{\textbf{2.37(0.03)}} & \textcolor{black}{\textbf{0.70(0.06)}} & \textcolor{black}{\textbf{2.92(0.02)}}  & \textcolor{black}{{0.93(0.02)}} & \textcolor{black}{\textbf{-1.22(0.00)}} & \textcolor{black}{\textbf{2.80(0.01)}} & \textcolor{black}{-4.45(0.02)} & \textcolor{black}{\textbf{2.76(0.01)}}\\
			\bottomrule
		\end{tabular}
	}	
\end{table*}

\begin{table*}[!h]
	\caption{RMSE values of 8 benchmark datasets. We report average and standard error over 20 runs.}
	\label{tab:uci_res_rmse}
	\centering
	\resizebox{0.9\textwidth}{!}{
		\begin{tabular}{ l c  cc cc cc cc}
			\toprule
			& & Boston & Energy & Concrete & Wine Red & Kin8nm & Power & Naval & Protein \\
			\cmidrule(lr){3-10}
			& $N$ & 506 & 768 & 1,030  & 1,599 & 8,192 & 9,568 & 11,934 & 45,730  \\
			& $D$ & 13 & 8 & 8 & 22 & 8 & 4 & 26& 9  \\
			\midrule
			
			Dropout \cite{dropout_gal} && \textbf{2.97(0.19)} & 1.66(0.04) & 5.23(0.12) &\textbf{0.62(0.01)}& {0.10(0.00)} &
			\textbf{4.02(0.04)}& \textbf{0.01(0.00)}&\textbf{4.36(0.01)}\\
			DVI \cite{wu2018deterministic}&& - & - & - & - & - & - & - & -\\
			 \midrule
			NSDE-MC \cite{adjoint_sde}\\
			Integration Time: 2s && 3.97(0.14) & 2.68(0.07) & 6.14(0.10) & 0.66(0.01)&\textbf{0.08(0.00)} & 4.42(0.03)&\textbf{0.01(0.00)}& 4.67(0.01)\\
			Integration Time: 4s && 3.65(0.15) & 2.38(0.19) & 6.02(0.08) & 0.66(0.01)&\textbf{0.08(0.00)} & 4.45(0.03)&\textbf{0.01(0.00)}& 4.66(0.01)\\
			\textcolor{black}{Integration Time: 8s} && \textcolor{black}{3.84(0.19)} & \textcolor{black}{\textbf{0.73(0.08)}} & \textcolor{black}{6.18(0.13)} & \textcolor{black}{0.68(0.01)}& \textcolor{black}{{0.09(0.00)}} & \textcolor{black}{4.56(0.03)}& \textcolor{black}{\textbf{0.01(0.00)}}& \textcolor{black}{4.77(0.02)}\\
			 \midrule
			NSDE-Cubature (Ours, Ablation)  \\
			Integration Time: 2s && 4.34(0.16) & 2.08(0.06)  & 6.49(0.10) & 0.64(0.00)&{0.10(0.00)} & 4.15(0.03)&\textbf{0.01(0.00)}& 4.51(0.02)\\ 
			Integration Time: 4s && 3.80(0.15) & 1.37(0.03)  & 6.08(0.09) & 0.64(0.00)&{0.09(0.00)} & 4.17(0.03)&\textbf{0.01(0.00)}& 4.44(0.02)\\
			\textcolor{black}{Integration Time: 8s} && \textcolor{black}{3.49(0.15) } & \textcolor{black}{1.12(0.04)}  & \textcolor{black}{5.99(0.09)} & \textcolor{black}{0.64(0.00)}& \textcolor{black}{{0.09(0.00)}} & \textcolor{black}{4.16(0.03)}& \textcolor{black}{\textbf{0.01(0.00)}}& \textcolor{black}{4.44(0.02)}\\
			  \midrule
			  NSDE-BMM (Ours, Proposed)\\
			  Integration Time: 2s && 3.41(0.13) & 2.23(0.09) & 5.47(0.10) & 0.64(0.01)&\textbf{0.08(0.00)} & 4.07(0.03) & \textbf{0.01(0.00)}& 4.63(0.01)\\
			  Integration Time: 4s && {3.17(0.14)} & {1.40(0.18)} & {5.26(0.10)} & {0.64(0.00)}&\textbf{0.08(0.00)} & 4.11(0.04) & \textbf{0.01(0.00)}& 4.48(0.00)\\
			  \textcolor{black}{Integration Time: 8s} && \textcolor{black}{{3.26(0.15)}} & \textcolor{black}{{0.87(0.13)}} & \textcolor{black}{\textbf{5.12(0.09)}} & \textcolor{black}{\textbf{0.63(0.00)}}& \textcolor{black}{\textbf{0.08(0.00)}} & \textcolor{black}{{4.07(0.03)}} & \textcolor{black}{\textbf{0.01(0.00)}} & \textcolor{black}{4.45(0.00)}\\
			\bottomrule
		\end{tabular}
}	
\end{table*}

\begin{figure*} [!h]
        \centering
        \begin{subfigure}[b]{0.245\textwidth}
            \centering
            \includegraphics[width=\textwidth]{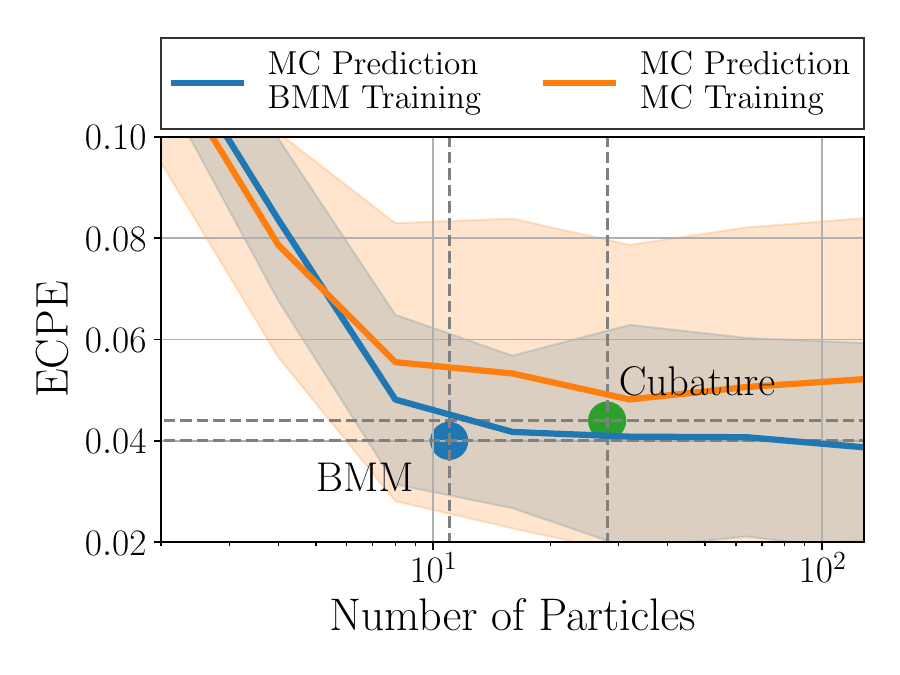}
            \caption[Boston]%
            {{\small Boston}}    
        \end{subfigure}
        \hfill
        \begin{subfigure}[b]{0.245\textwidth}  
            \centering 
            \includegraphics[width=\textwidth]{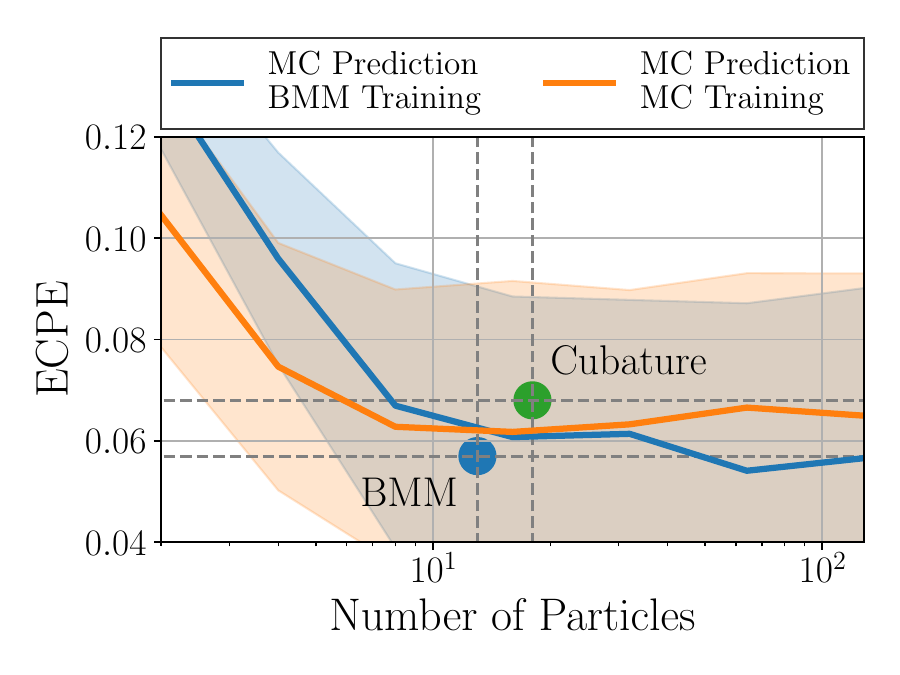}
            \caption[Energy]%
            {{\small Energy}}    
        \end{subfigure}
        \hfill
        \begin{subfigure}[b]{0.245\textwidth}   
            \centering 
            \includegraphics[width=\textwidth]{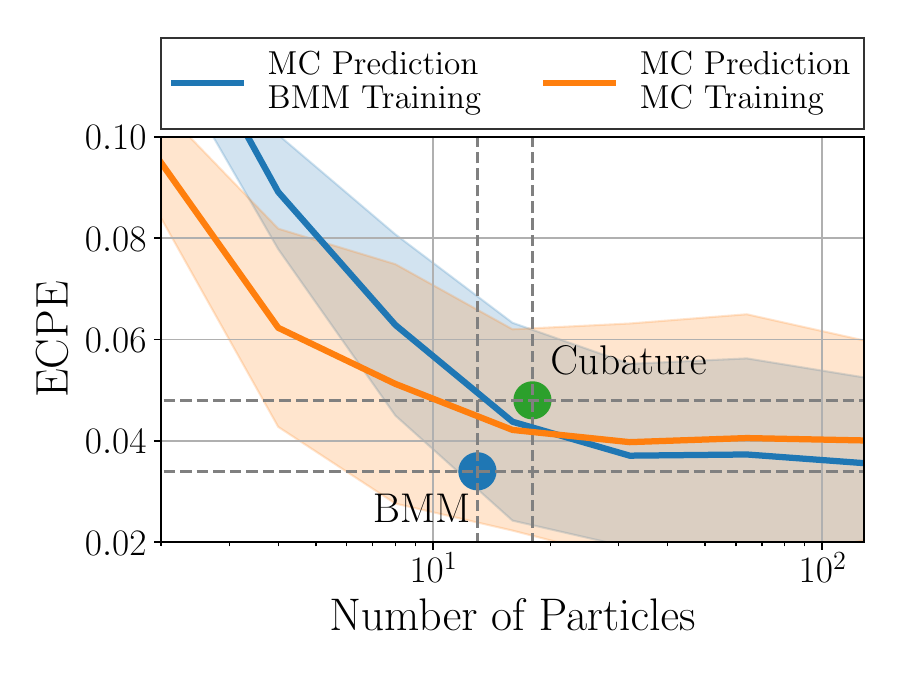}
            \caption[Concrete]%
            {{\small Concrete }}    
        \end{subfigure}
        \hfill
        \begin{subfigure}[b]{0.245\textwidth}   
            \centering 
            \includegraphics[width=\textwidth]{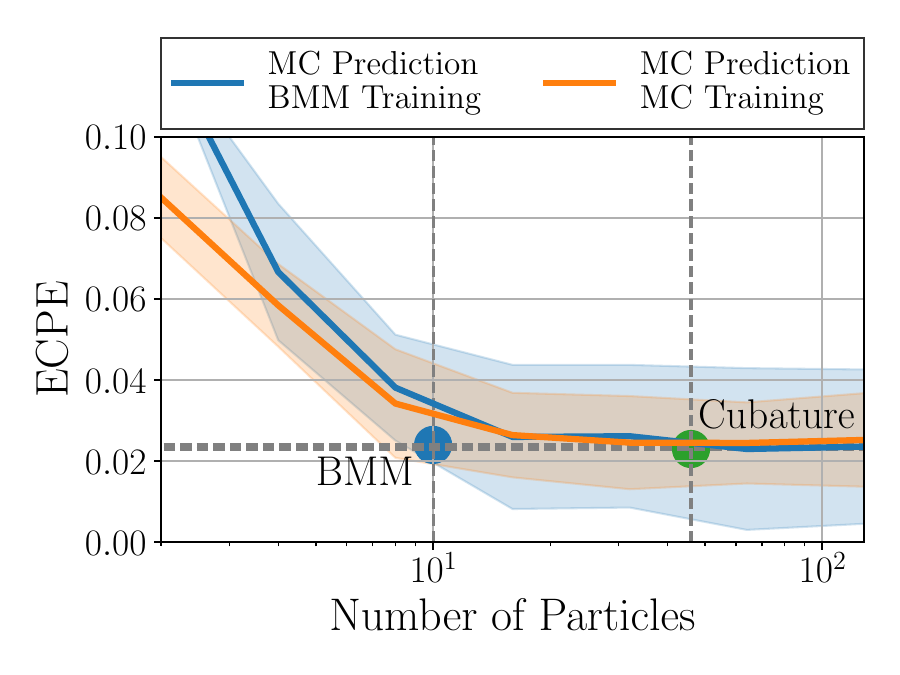}
            \caption[Wine Red]
            {{\small Wine Red }}    
        \end{subfigure}
        \centering
        \begin{subfigure}[b]{0.245\textwidth}
            \centering
            \includegraphics[width=\textwidth]{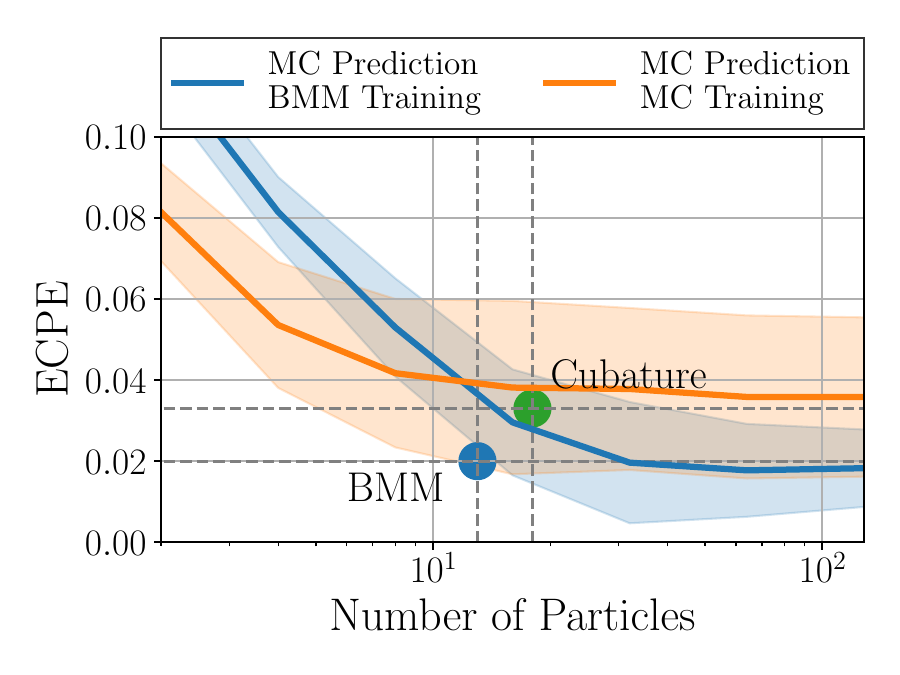}
            \caption[Kin8nm]%
            {{\small Kin8nm}}    
        \end{subfigure}
        \hfill
        \begin{subfigure}[b]{0.245\textwidth}  
            \centering 
            \includegraphics[width=\textwidth]{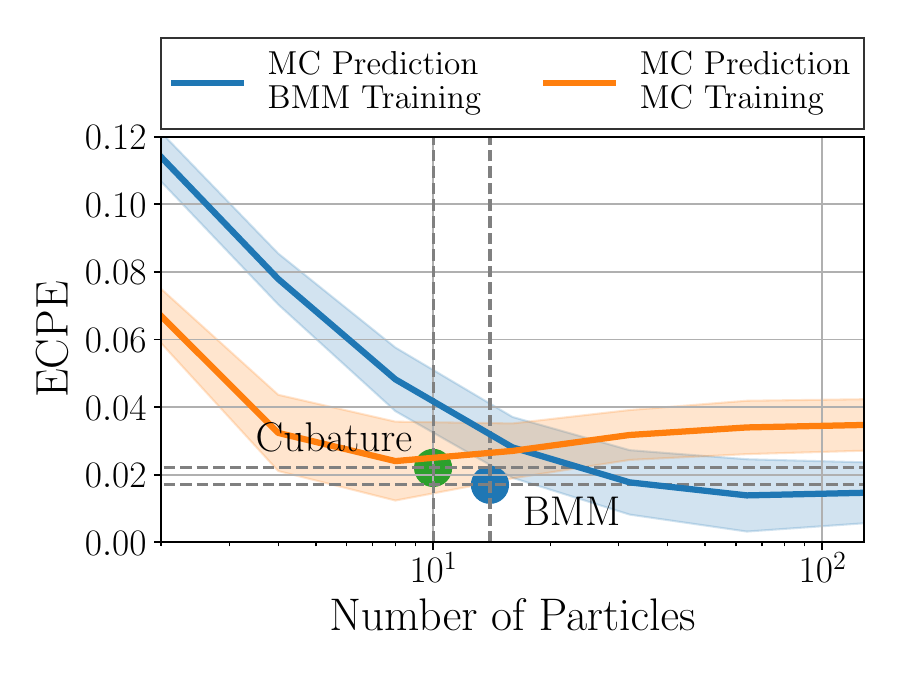}
            \caption[Power]%
            {{\small Power}}    
        \end{subfigure}
        \hfill
        \begin{subfigure}[b]{0.245\textwidth}   
            \centering 
            \includegraphics[width=\textwidth]{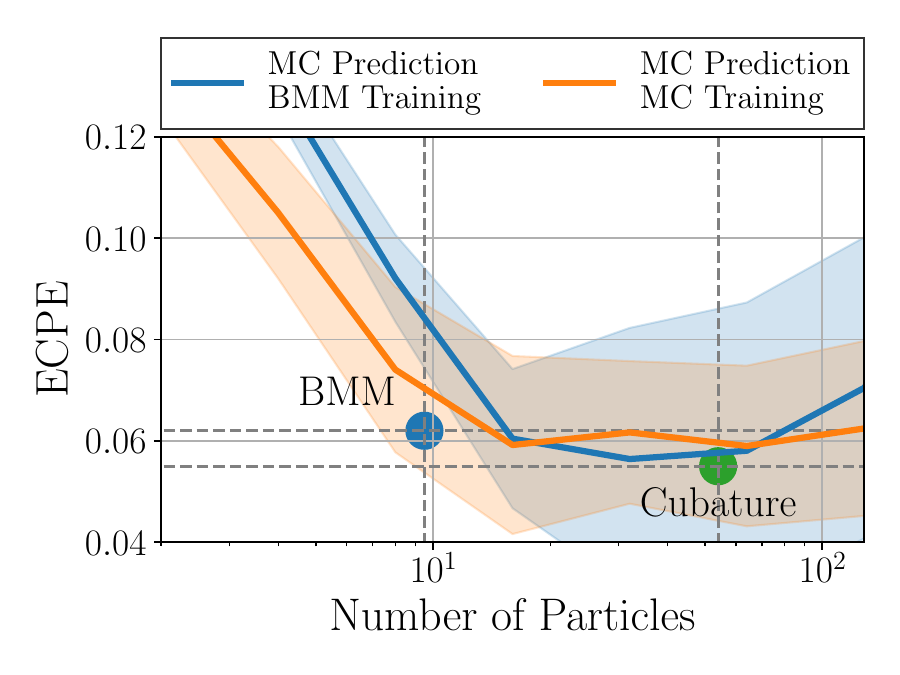}
            \caption[Naval]%
            {{\small Naval }}    
        \end{subfigure}
        \hfill
        \begin{subfigure}[b]{0.245\textwidth}   
            \centering 
            \includegraphics[width=\textwidth]{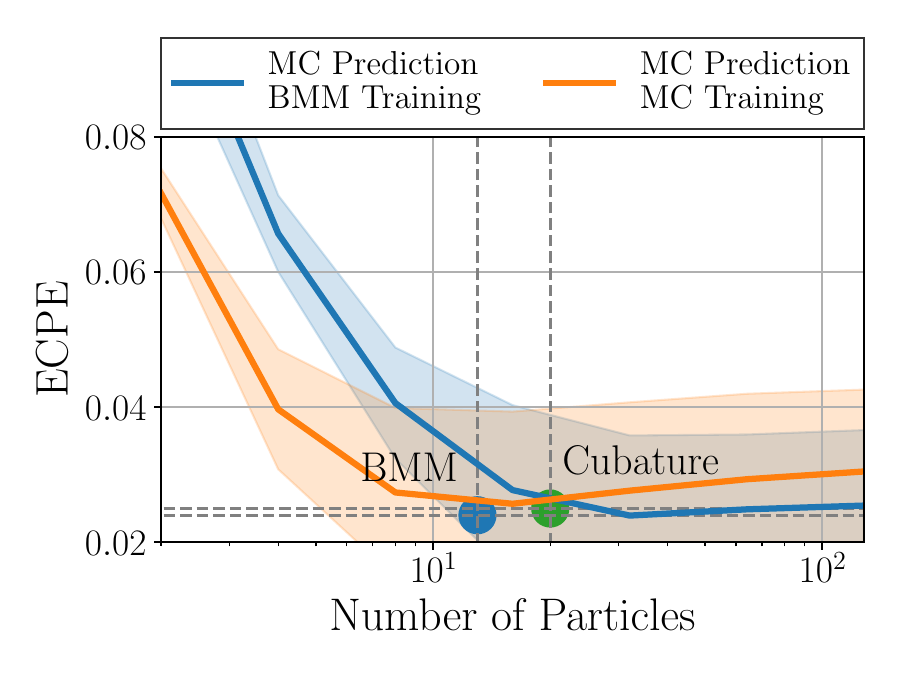}
            \caption[Protein]%
            {{\small Protein }}    
        \end{subfigure}%
        \caption[calibration-cost-uci]
        { Cost-benefit analysis of calibration with different methods on the regression task for a flow time of \textcolor{black}{8} seconds. One particle is equal to one MC simulation along a trajectory.
        BMM is our proposed method, Cubature is our method that uses cubature for VMM, the orange line is training and prediction with MC sampling, and the blue line is training with our method and prediction with MC sampling. 
        We show mean and standard deviation over 10 runs. } 
        \label{fig:uci_calibration}
    \end{figure*}

NSDEs  can be used as a general-purpose layer for neural networks, just as neural ODEs \cite{neural_ode}. 
\textcolor{black}{Similarly, \cite{nsde_raginsky} proposed latent NSDEs as a layer and derived a continuous time evidence lower bound, which was maximized during training by taking samples.
In contrast to \cite{nsde_raginsky}, we overcome the need of taking samples, propose an alternative training objective based on maximum likelihood estimation, and show for the first time the applicability of NSDEs as layers on a real world dataset.}
We demonstrate its application on a regression task.
Given an input $x \in \mathbb{R}^D$, we aim to predict the target value $y \in \mathbb{R}$.  
We propagate the input $x=x(t_0)$ through the NSDE for a varying flow time $t_1$ and obtain as a result $x(t_1)$, from which we predict $y$. The mapping from $x(t_0)$ to $x(t_1)$ can be interpreted as a continuous depth layer. 
\textcolor{black}{We formulate the  probability of observing the target value conditioned on the input as
\begin{equation}
    p(y|x(t_0), \theta, \phi, \psi) \!\!=\!\! \int \!\! p(x(t_1) | x(t_0), \theta, \phi) p(y | x(t_1), \psi) d x(t_1),
\end{equation}
where $p(x(t_1) | x(t_0), \theta, \phi)$ is determined by the underlying NSDE with parameters $\theta$ and $\phi$ and can be evaluated using our BMM scheme. We further assume that the mapping from $x(t_1)$ to $y$ is linear, which allows us to analytically marginalize out $x(t_1)$ from the above expression.
We learn the parameters $\theta, \phi, \psi$  by \textit{Maximum Likelihood Estimation } (MLE) 
\begin{equation}
\hat{\theta}, \hat{\phi}, \hat{\psi} = \underset{\theta, \phi, \psi}{\text{argmax}}  ~ \mathbb{E} \left[  p(y|x(t_0), \theta, \phi, \psi)\right].
\label{eq:mle}
\end{equation}}
In this experiment, we vary the time $t_1$ between \textcolor{black}{$2$} and \textcolor{black}{$8$} seconds and choose $dt=0.5$ seconds. 

\subsubsection{Datasets}
We use eight UCI datasets, which have varying input dimensionality $D$ and size $N$. 
These datasets  \textcolor{black}{are commonplace used in prior art for benchmarking stochastic models such as neural networks \cite{dropout_gal, pbp} as well as Gaussian Processes \cite{diffgp, double_vi, lindinger2020beyond}}. 
The datasets are available  
\href{http://archive.ics.uci.edu/ml}{here}. We use the experimental setup as  defined in \cite{pbp}. \textcolor{black}{We use 20 random splits, where $90\%$ of the data are used for training and $10\%$ are used for testing.} 
We use for all datasets a batch size of 32.

\subsubsection{Baselines} We benchmark our method (BMM) against different NSDE variants as well as against commonly used neural regression models.
\textcolor{black}{Our NSDE variants have $103D+51$ parameters.
We use a neural drift function with one hidden layer of size 40 with $81D+40$ parameters and a neural diffusion function with one hidden layer of size 10 with $21D+10$ parameters.
After propagating an input for a defined time horizon  we use an affine transformation with $D+1$ parameters in order to map the NSDE prediction to the regression target. 
The dropout rate has been adjusted for each dataset seperately.}

\textit{(i) NSDE-MC} \cite{adjoint_sde}: Monte Carlo sampling-based prediction with NSDE. We use for NSDE-MC $2(2D+1)$ particles, as it equals the number of function evaluations for NSDE-Cubature.

\textit{(ii) Cubature:} A variant of ours using cubature in place of VMM to approximate $\mu_{k+1}$ in Eq. \ref{eq:moment-matching-mean} and covariance $\Sigma_{k+1}$ in Eq. \ref{eq:moment-matching-covariance}. 

\textit{(iii) DVI} \cite{wu2018deterministic}: 
\textcolor{black}{
DVI is an inference technique for Bayesian neural nets. In this model class,
one arrives at a probabilistic model by allowing for uncertainty over the 
weights. The evidence lower bound is approximated in a deterministic manner by applying moment
matching rules.
DVI has $100D + 202$ parameters. This makes the NSDE variants slightly more parameter efficient as $D\leq 50$ in all experiments.
DVI has the computational cost of $\mathcal{O}(H^3)$, where $H$ is the hidden layer width. For comparison, our method has the computational cost of $\mathcal{O}(KH^3)$ as described in Sec. \ref{subsec:computational_cost}.}

\textit{(iv) Dropout} \cite{dropout_gal}: This method introduces stochasticity by applying a Bernoulli distributed masking scheme in the affine layers. This method has the least parameters: $50D + 101$.
\textcolor{black}{Dropout has the computational cost of $\mathcal{O}(SH^2)$, where $S$ corresponds to the number of MC evaluations.}

\textcolor{black}{Dropout and DVI are strongly linked, as Dropout can be interpreted as the sampling-based version of DVI \cite{variational_dropout}. 
In consequence, given a limited computational budget, Dropout can lead to high gradient variance during training, which can result in deteriorated solutions.
Both DVI and Dropout assume a probabilistic model by injecting noise over the neural net weights, which corresponds to the noise due to the lack of knowledge.
In contrast, our method is applicable to neural stochastic differential equations which is a model family for continuous-time stochastic processes. Given an initial value $x(t_0)$, the solution at time point $t_n$ is characterized by the probability distribution $p(x(t_n) | x(t_0))$.
Our proposed model captures input noise, which corresponds to irreducible noise factors of the data generating process, and parameter noise.
Input noise is modeled  via the diffusion term and parameter noise via Dropout. As our model captures both noise sources in contrast to Dropout and DVI, we expect our model to quantify uncertainty more accurately in terms of lower NLL. 
Furthermore, when our method is used as a continuous layer for regression tasks, we can adjust the model capacity by increasing the flow time while keeping the number of weights constant. Contrarily, the model capacity of DVI as well as of Dropout is fixed by the architecture. 
In consequence, DVI and Dropout have the same disadvantages compared to our method from a modeling perspective, i.e. they are less parameter efficient, do not model input noise, and cannot be used out-of-the-box for continuous-time stochastic systems.}

\subsubsection{Results}
We report NLL in table \ref{tab:uci_res_nll} and RMSE in table \ref{tab:uci_res_rmse}. 
Despite the fact that NSDEs are dynamical systems, they can achieve competitive results as a layer compared to standard neural networks.
NSDE BMM shows decreasing RMSE and NLL with increasing flow time since the expressiveness of the NSDE also increases. 
\textcolor{black}{
Our method BMM has lower NLL than  DVI in five datasets and the same NLL in one dataset. We account the increased performance of our model to its higher parameter efficiency and capability of modeling both noise sources, i.e. parameter and input noise.
In contrast to our proposed NSDE variant, increasing the integration time leads in some datasets to degraded predictive performance for the MC and cubature NSDE variants.
We attribute the improved prediction accuracy of BMM over cubature to its reduced approximation error, as demonstrated in Fig. \ref{fig:integration}.}
We observe in Fig. \ref{fig:uci_calibration} the ECPE of our method BMM to decrease or to be on par with NSDE-MC. When using our method only for training and instead performing testing by drawing samples, we observe the same calibration levels as directly testing with our method. This indicates a tight approximation of BMM, since the same uncertainty calibration is reached as the infinite particle limit at a lower computational cost.

\subsection{Time Series Classification}
\label{sec:time_classification}
In time series classification, we aim to predict the label $y \in \{ 1, 2, \ldots, K\}$ after observing the time series $U=\{u(t_n)\}_{n=0}^{N-1}$ for $N$ time steps. 
We treat the observations $u(t_n)$ as inputs to a latent NSDE with an input dependent transition kernel $p(x(t_{n+1})|x(t_n), u(t_n), \theta, \phi)$.  
The probability of observing the label $y$ depending on the inputs $U$ is 
\begin{align}
    p(y|U, \theta, \phi,\psi) = \int &p(y|x(t_N), \psi)  p(x(t_N) | U, \theta, \phi) d  x(t_N),
\end{align}
with the marginal distribution 
\begin{align}
    p(x(t_N) | U, \theta, \phi)  = \int \prod_{n=0}^{N-1}  &p(x(t_{n+1})| x(t_n), u(t_n), \theta, \phi) \\
                                             &\times p(x(t_0))  d x(t_0), \ldots, x(t_{N-1}) \nonumber.
\end{align}
As many time series are equally spaced, i.e. $\forall n: t_{n+1}-t_n= \text{const.} $, a reasonable choice is to set the time step as $\Delta t = t_{n+1} -t_n$. The marginal distribution becomes 
\begin{align}
    p(x(t_N) | U,\! \theta,\! \phi) \approx \!\! \int \!\prod_{n=0}^{N-1} \!\! &\mathcal{N}  (x_{n+1}  | m_{n+1}(x_n,\! u_n), S_{n+1}(x_n,\! u_n) \!) \nonumber \\ 
                                                  &\times p(x_0)  d x_0, \ldots, x_{N-1}, 
\end{align}
where $x_n = x(t_n)$, $u_n = u(t_n)$, $m_{n+1}(x_n) := x_n +  f_{\theta}(x_n, u_n)\Delta t$ and $S_{n+1}(x_n, u_n) := {L}_{\phi}{L}_{\phi}^T(x_{n}, u_n)  \Delta t$. 
Our algorithm BMM allows us to efficiently approximate the marginal distribution $p(x(t_N) | U, \theta, \phi)$.
The conditional distribution $p(y|x(t_N), \psi)$ is modeled as a linear layer with parameters $\psi$ followed by softmax in order to map $x(t_N)$ to the distribution over class labels. Following \cite{wu2018deterministic}, the output moments of the softmax layer can be closely approximated, which allows us to compute the target distribution $p(y|U, \theta, \phi, \psi)$ via moment matching. 
The parameters $\theta, \phi, \psi$ are inferred by MLE as in Eq. \ref{eq:mle}. 
\
\subsubsection{Datasets} We benchmark on two datasets: 
\\
\indent
\textit{ (i) MNIST.} This dataset consists of 60k training and 10k testing images of single digits between 0 and 9. Each image has size $28 \times 28$. We treat the images as time series with length 28 and dimensionality 28. We use a batch size of 10.
\\
\indent
\textit{ (ii) IMDB.} This dataset consists of 37500 training and 12500 testing sequences. Each sequence corresponds to a movie review and has varying length between 2 and 652 words. \textcolor{black}{We follow the  
{tutorial}\footnotemark[1] for selecting hyperparameters (e.g. batch size, word dictionary size) since it
(i) is easy to reproduce, and 
(ii) has already been applied in the existing literature \cite{kandemir2022evidential}, where it was shown to be capable to highlight differences between models and inference strategies.
} 
We generate for each run a word dictionary of size 1000 and omit sequences that are longer than 500 words, which are in total 3 omitted sequences.  \textcolor{black}{We use a batch size of 50.}
\
\subsubsection{Baselines} 
\label{subsubsec:time_classification_baselines}
We use the same NSDE variants as in the previous section, i.e. NSDE-MC and NSDE-Cubature.
We use drift neural networks with two hidden layers and and diffusion neural networks with one hidden layer. We choose a hidden layer size of 50 and a latent dimensionality of 32. 
Additionally, we compare against two competitive time series classification models:
\\
\indent
\textit{(i) LSTM}: We use a stacked LSTM, consisting of two LSTMs with a hidden layer size of 64. 
\\
\indent
\textit{(ii) Transformer}: Our Transformer architecture follows the {original work}\footnotemark[2]. We use 2 Transformer encoder layers with hidden size 64 and 2 attention heads. 
\\
The NSDE/LSTM/Transformer models have 11k/62k/53k parameters for the MNIST experiment and 40k/130k/120k parameters for the IMDB experiment. 
\
\footnotetext[1]{\url{https://www.kaggle.com/code/arunmohan003/sentiment-analysis-using-lstm-pytorch/notebook}} 
\footnotetext[2]{\url{https://pytorch.org/tutorials/beginner/transformer_tutorial.html}}
\subsubsection{Results}
We present time series classification results in table \ref{tab:results_time_classification}. 
We find our model NSDE-BMM to achieve the same level of performance as LSTM despite having less parameters and not being tailored towards modeling of long-term effects, demonstrating the applicability of our model as a general purpose tool. 
Our proposed model NSDE-BMM outperforms NSDE-MC and NSDE-Cubature. 
For the IMDB dataset, NSDE-Cubature cannot be trained with a GPU with 40GB due  to an \textit{Out of Memory} (OOM) error.
We observe the Transformer model to be outperformed by LSTM and NSDE-BMM due to possible overparameterization.  

\textcolor{black}{Our experimental results also hold when varying the hyperparameters; When rerunning the IMDB experiment with a word dictionary of size 10,000, a batch size of 250, and a simple (one layer) architecture with 32 neurons and applying
Dropout (0.25), we can increase the performance of all methods by up to 3\% without changing the relative ranking between the methods.}

\begin{table}[h!]
\centering
	\caption{\textcolor{black}{Results for different time series classification tasks. 
	We provide average  and standard error over 10 runs. ECE is reported in percent.}}
	\label{tab:results_time_classification}
	\resizebox{\columnwidth}{!}{
	\begin{tabular}{l ccc|ccc}
		\toprule
		& \multicolumn{3}{c}{MNIST} & \multicolumn{3}{c}{IMDB}\\
		\cmidrule{2-7}
		& ACC & NLL & ECE & ACC & NLL & ECE\\
		\cmidrule{2-7}
		LSTM         & 98.19(0.05) & 0.06(0.00) & 2.16(0.07) & 85.61(0.12) & 0.34(0.00) & 15.17(0.11)\\
		Transformer  & 95.87(0.21) & 0.13(0.01) & 5.12(0.20) & 82.33(0.41) & 0.40(0.01) & 17.59(0.28)\\
        \midrule
		NSDE-MC \cite{adjoint_sde}                   & 81.60(0.53) & 0.58(0.01) &  20.00(0.30) & 84.92(0.07) & 0.35(0.00) & 16.72(0.09)\\
		NSDE-Cubature  & \multirow{ 2}{*}{95.16(0.81)} & \multirow{ 2}{*}{0.19(0.03)} & \multirow{ 2}{*}{9.19(1.07)} & \multirow{ 2}{*}{OOM} & \multirow{ 2}{*}{OOM} & \multirow{ 2}{*}{OOM}\\
		(Ours, Ablation) \\
	    NSDE-BMM & \multirow{ 2}{*}{98.11(0.16)} & \multirow{ 2}{*}{0.06(0.03)} & \multirow{ 2}{*}{2.52(0.09)} & \multirow{ 2}{*}{85.87(0.16)} & \multirow{ 2}{*}{0.33(0.00)} & \multirow{ 2}{*}{15.48(0.12)}\\
	    (Ours, Proposed)\\
		\bottomrule
		\end{tabular}
		}
\end{table}

\subsection{Time Series Modeling}
\label{sec:time_modeling}
As NSDEs are describing dynamical systems they are a natural choice for time series modeling. 
\textcolor{black}{Together with the EM
discretization we may approximate the joint distribution of an equally spaced time series $X=\{x(t_n)\}_{n=0}^N=\{x_n\}_{n=0}^N$ governed by the NSDE in Eq. \ref{eq:nsde}, which is  observed on $N+1$ time points, as
\begin{equation}
    p(X| \theta, \phi) \approx  p(x_0) \prod_{n=0}^{N-1} \mathcal{N}  (x_{n+1}  | m_{n+1}(x_n), S_{n+1}(x_n) ).
    \label{eq:mle_time}
\end{equation}}
Given a set of time series, the parameters $\theta, \phi$ can be inferred by \textcolor{black}{MLE as in Eq. \ref{eq:mle}}. This training objective requires one-step predictions and no Monte Carlo approximations. We instead propose multi-step training with our method BMM in order to improve long-term predictions.
For multi-step training we need to approximate the transition kernel \textcolor{black}{$p(x_{n+j}|x_n, \theta, \phi)$} for some $j > 0$. 
During multi-step training, we maximize the augmented likelihood objective 
\textcolor{black}{
\begin{equation}
    p(X^j| \theta, \phi)\! =\!  p(x_0)\! \prod_{n=0}^{\left\lfloor \frac{N-1}{j} \right\rfloor}\! p(x_{nj+j}|x_{nj}, \theta, \phi),
\end{equation}
for $j\in \{1, \ldots, N-1\}$.} Above $X^j$ is the augmented time series, where only every $j-$th value is observed.
We provide a comparison to one-step training in the following experiments.

\subsubsection{Datasets} We benchmark on three datasets: 

\textit{ (i) Lotka-Volterra.}
We choose  stochastic Lotka-Volterra equations as in \cite{Abbatietal19}
 \begin{equation*}
d  x_t=\begin{bmatrix} 2 x_{t,1} -  x_{t,1} x_{t,2}\\  x_{t,1} x_{t,2} - 4 x_{t,2}\end{bmatrix}dt  + \sqrt{\begin{bmatrix} 0.05 & 0.03\\ 0.03 & 0.09\end{bmatrix}} d  w_t.
\end{equation*}
We generate 128 paths using Euler-Maruyama discretization with a small step size of $dt=10^{-5}$ seconds. Afterwards, we coarsen the dataset such that 200 equally spaced observations between $0-10$ seconds remain.  
First 100 observations are used for training and the remaining 100 observations for testing. We use a batch size of 16 and a prediction horizon of $10$ steps. 

\textit{(ii) Beijing  Air Quality.} The atmospheric air-quality dataset from Beijing \cite{air_quality} consists of hourly measures over the period 2014-2016 at three different locations. 
The air quality is characterized by 10  different features at each location. 
Including the timestamp, we obtain in total 34 features.   
We follow \cite{wishart_sde} for designing the experimental setup. The first two years are used for training and we test on the first 48\,hours in the year 2016. We use a batch size of 16 and a prediction horizon of $10$ steps for training.
The dataset is available here\footnotemark[3].
\footnotetext[3]{\url{https://archive.ics.uci.edu/ml/datasets/Beijing+Multi-Site+Air-Quality+Data}}

\textit{(iii) 3-DOF-Robot.}
The 3-DOF-Robot dataset \cite{robot_arm} consists of multiple trajectories with length 14000,  3 input and 9 output dimensions, recorded with a sampling rate of 1kHz. 
The dataset was recorded at two different operating modes: (i) 50 recordings of low frequency oscillations, and (ii) 50 recordings of high frequency. We train on the first 38 trajectories and validate on the next 3 trajectories using the low frequency recordings. 
We use as a test set the final 9 low frequency trajectories (IID), and the final 9 high frequency recordings (Transfer).
We use a batch size of 16 and a prediction horizon of $16$ steps for training.
The dataset is available
here\footnotemark[4].
\footnotetext[4]{\url{https://owncloud.tuebingen.mpg.de/index.php/s/3THSfyBgFrYykPc?path=\%2F}}

\subsubsection{Baselines} 
We use the same NSDE variants as in the previous section, i.e. NSDE-MC and NSDE-Cubature, and train them on multi-step predictions.
We use drift neural networks with two hidden layers and diffusion neural networks with one hidden layer. The hidden layer size is 100 for Beijing  Air Quality and 50 for the other datasets.
\textcolor{black}{Similarly, as in the previous section we use  LSTM/Transformer models and modify them to predict a deterministic or stochastic output. The stochastic output of both models follows a Gaussian distribution for which we predict the mean and  variance at each prediction step. We train LSTM/Transformer models on one step predictions as we found multi-step training to produce deteriorated results due to high variance.}
Additionally we compare against:

\textit{(i) NSDE-One-Step}: A NSDE trained on one-step predictions.  The training objective is  deterministically tractable (see Eq. \ref{eq:em_normal}). Testing is done with Monte Carlo rollouts. 

\textit{(ii) NODE} \cite{neural_ode}: Neural ODE, hence NSDE without diffusion. As this method has no stochastic component, we report only MSE.

\textit{(iii) diffWGP} \cite{wishart_sde}: A SDE with drift and diffusion modeled as the predictive mean and covariance of a GP, the state of the art of differential equation modeling with GPs.

\textcolor{black}{
The NSDE/NODE/LSTM/Transformer models have 3k/3k/42k/122k parameters for the Lotka-Volterra, 24k/17k/61k/132k parameters for the Beijing  Air Quality, and 5k/4k/57k/220k parameters for the 3-DOF-Robot dataset.}

\subsubsection{Results}
 As shown in Tab. \ref{tab:results}, BMM outperforms all baselines in all datasets with respect to both NLL and MSE, with the only exception of NLL in the 3-DOF-Robot (IID) dataset \textcolor{black}{and MSE in the weather dataset}  in which it performs second best. BMM proves to make more accurate predictions than both NSDE-MC and diffWGP probably due  to the improved stability of the training process thanks to its deterministic objective. This outcome is despite the fact that diffWGP models the diffusion as a Wishart process, while BMM uses a diagonal diffusion function.
 Our results indicate that using a one-step training objective may hinder learning of long term relations, as NSDE-One-Step performs worse than  NSDE-Cubature and our method in all datasets. 
 \textcolor{black}{We find that our model outperforms LSTM/Transformer on stochastic time-series modeling tasks. While we cannot rule out other
causes, this might be due to the different objectives when designing
the architectures: NSDEs are targeted towards
stochastic dynamical systems, while LSTM/ Transformer architectures aim to capture long-term effects that are more frequent in language modeling tasks than in the data sets used in this experiment. We observe in all datasets that changing the Transformer/LSTM architecture from a deterministic to a stochastic output results in higher MSE. A deterministic multi-step training objective for these two methods, as proposed for NSDEs in this work, seems to be a promising future research direction.}

\begin{figure*} [h]
        \centering
        \begin{subfigure}[b]{0.245\textwidth}
            \centering
            \includegraphics[width=\textwidth]{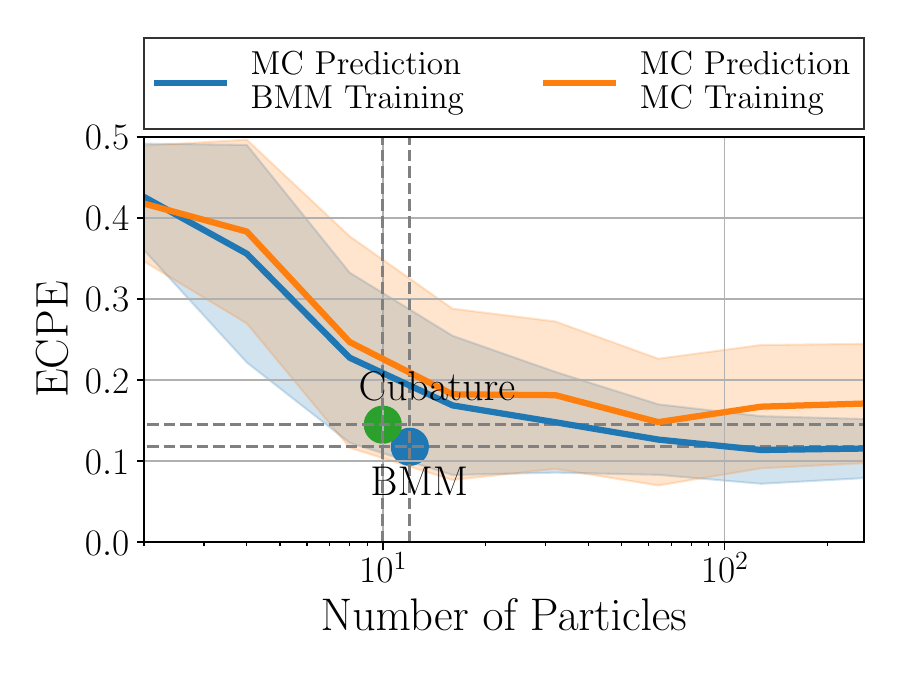}
            \caption[Lotka-Volterra]%
            {{\small Lotka-Volterra}}    
        \end{subfigure}
        \hfill
        \begin{subfigure}[b]{0.245\textwidth}  
            \centering 
            \includegraphics[width=\textwidth]{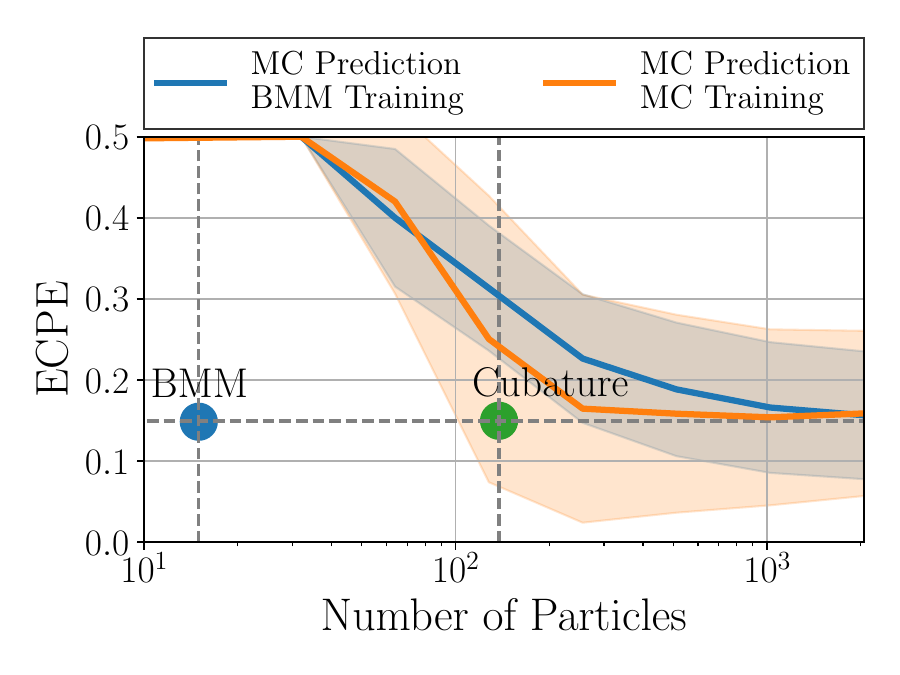}
            \caption[Air Quality\newline]%
            {{\small Air Quality}}    
        \end{subfigure}
        \hfill
        \begin{subfigure}[b]{0.245\textwidth}   
            \centering 
            \includegraphics[width=\textwidth]{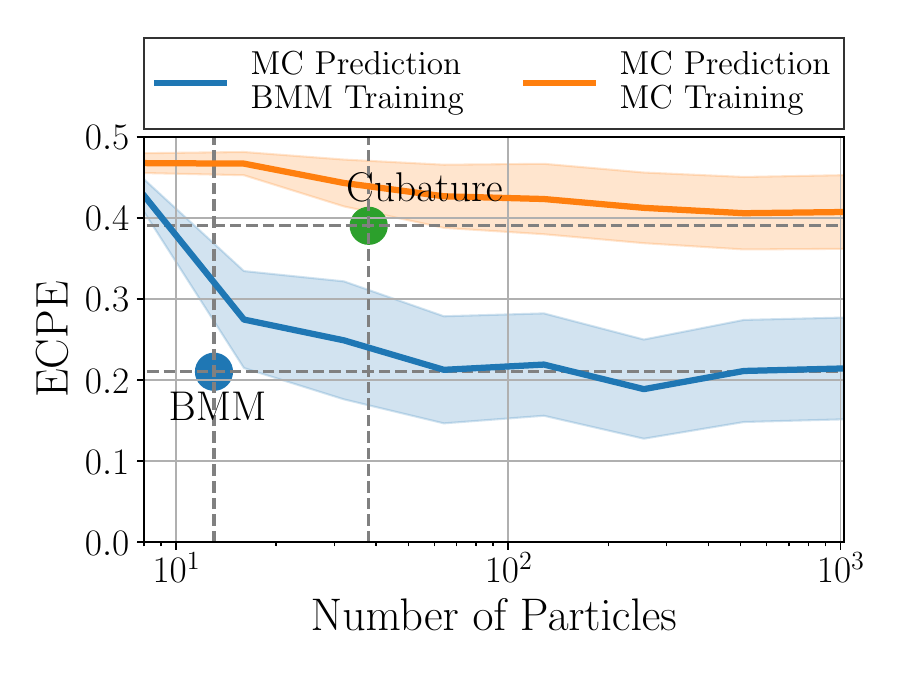}
            \caption[3-DOF-Robot IID]%
            {{\small 3-DOF-Robot, IID }}    
        \end{subfigure}
        \hfill
        \begin{subfigure}[b]{0.245\textwidth}   
            \centering 
            \includegraphics[width=\textwidth]{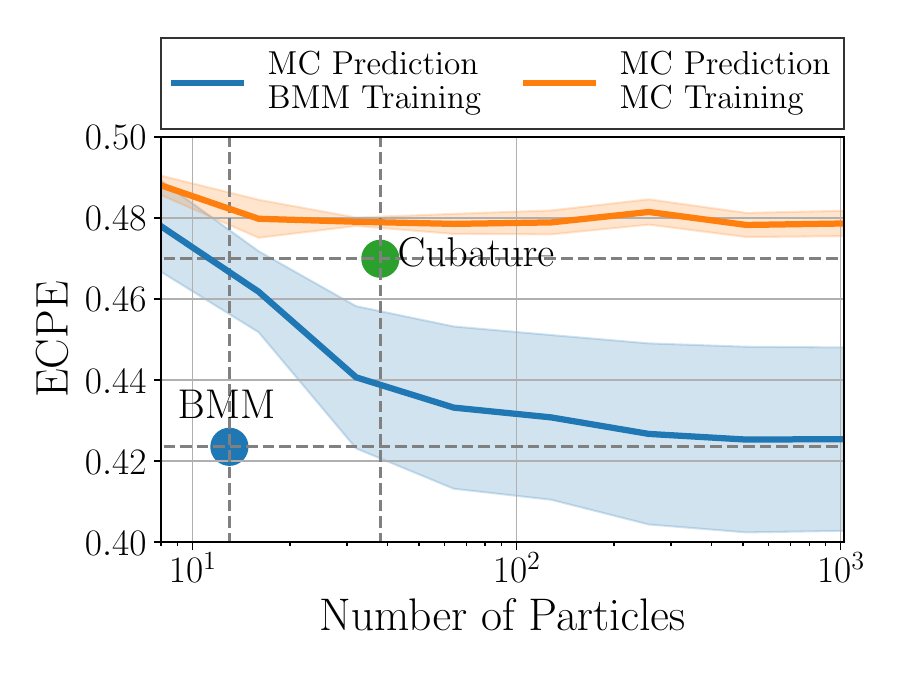}
            \caption[3-DOF-Robot Transfer]%
            {{\small 3-DOF-Robot, Transfer }}    
        \end{subfigure}
        \caption[calibration-cost]
        {{ Cost-benefit analysis of calibration with different methods.} One particle is equal to one MC simulation along a trajectory.
        BMM is our proposed method, Cubature is our method that uses cubature for VMM, the orange line is training and prediction with MC sampling, and the blue line is training with our method and prediction with MC sampling. 
        We show mean and standard deviation over 10 runs. } 
        \label{fig:ecpe_time_series}
    \end{figure*}

Compared to the previous regression task, our method improves stronger on its baselines in terms of uncertainty calibration. Since small errors can potentially accumulate, predictions become increasingly challenging for longer horizons.
As shown in Fig. \ref{fig:ecpe_time_series} the BMM algorithm reaches a level of uncertainty calibration, which is prohibitively costly for MC sampling. In all three applications, MC sampling requires more than 50 roll-outs to match the ECPE, which our deterministic BMM provides. 
We observe BMM to bring smaller ECPE than cubature in three of the four plots at comparable or less computational cost.

\begin{table*}[h!]
\centering
	\caption{Forecasting results for different time series prediction task. 
	We provide average  and standard error over 10 runs.}
	\label{tab:results}
	
	\adjustbox{max width=0.9\textwidth}{
	\begin{tabular}{l c cc|cc|cccc}
				\toprule
				              &&     \multicolumn{2}{c|}{\multirow{2}{*}{Lotka-Volterra}} & \multicolumn{2}{c|}{\multirow{2}{*}{Air Quality}} & \multicolumn{4}{c}{3-DOF-Robot} \\
				              &&      & & & & \multicolumn{2}{c}{IID}  &\multicolumn{2}{c}{Transfer}  \\
				\cmidrule{3-10}
				& $D$ & \multicolumn{2}{c|}{2} & \multicolumn{2}{c|}{34} & \multicolumn{4}{c}{9}\\
				\cmidrule{3-10}
				
				 		 &                 & MSE 		        & NLL    & MSE 		   & NLL                          &  MSE  & NLL  & MSE & NLL   \\
				\midrule
	$\text{LSTM (Deterministic)}$          && 1.86(0.02)              & -      & 1.59(0.08) & - & 0.05(0.00) & - & 2.67(0.05) & -\\ 
    $\text{Transformer (Deterministic)}$   && 1.94(0.04)              & -      & \textbf{1.41(0.12)} & - & 0.21(0.02) & - & 3.52(0.17) & -\\ 
	$\text{LSTM (Stochastic)}$             && 2.41(0.09) & 5.32(0.14) & 2.37(0.04) & 55.79(0.43)& 0.43(0.03) & 9.49(0.23) & 3.03(0.06) & 131(6)\\ 
    $\text{Transformer (Stochastic)}$      && 2.19(0.06) & 5.03(0.06) & 2.49(0.05) & 55.43(0.44) & 0.31(0.01) & 10.07(0.41) & 3.62(0.16) & 144(9)\\ 
    $\text{NODE}$  \cite{neural_ode}      && 1.98(0.04) & -      & {1.85}({0.15}) & - & 0.04(0.00) & - & 3.78(0.12) & -\\ 
    $\text{diffWGP}$   \cite{wishart_sde} && - & -  & {2.39}({0.04})  & {46.92}\footnotemark[5]({0.75}) & -  & - & - & -  \\
    \midrule
    	NSDE-MC \cite{adjoint_sde}     && 2.07(0.11) & 4.95(0.44) & 1.90(0.07) & 43.84(0.75) & 0.04(0.00) & 21.69(1.75)  & 4.69(0.13) 
		& 1821(12) \\
	    NSDE-One-Step   (Ours, Ablation) && 2.44(0.14) & 5.38(0.27) & 1.75(0.09) & 62.53(0.96) & 0.05(0.00) & 23.40(1.34)  & 4.43(0.11) 
		& 1437(25)\\
	    NSDE-Cubature (Ours, Ablation) && 1.91(0.08) & 4.84(0.19) &{1.45}({0.04}) & 37.84(0.39) & 0.04(0.00) & \textbf{5.65(0.33)}  & 4.18(0.13) 
		&  352(27)\\
		NSDE-BMM (Ours, Proposed) && \textbf{1.75(0.03)}& \textbf{4.35(0.15)}& 1.44(0.10) & \textbf{34.30(0.50)} & \textbf{0.03(0.01)} & {7.27}({1.31}) & \textbf{2.31(0.08)} 
		& \textbf{102(12)}\\
		\bottomrule
		\end{tabular}
		}
\end{table*}

\begin{figure}[h]
	\begin{subfigure}[t]{0.5\columnwidth}
	\centering
	\includegraphics[width=\columnwidth]{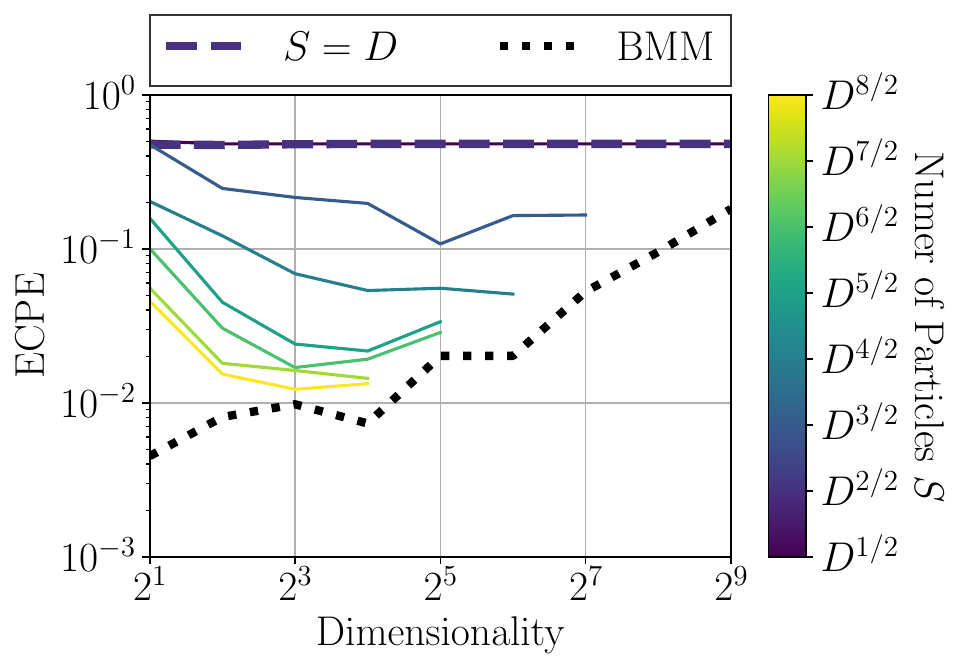}
	\caption{ECPE}
	\label{fig:var_dim_ecpe}
	\end{subfigure}
	\hfill
	\begin{subfigure}[t]{0.5\columnwidth}
	\centering
	\includegraphics[width=\columnwidth]{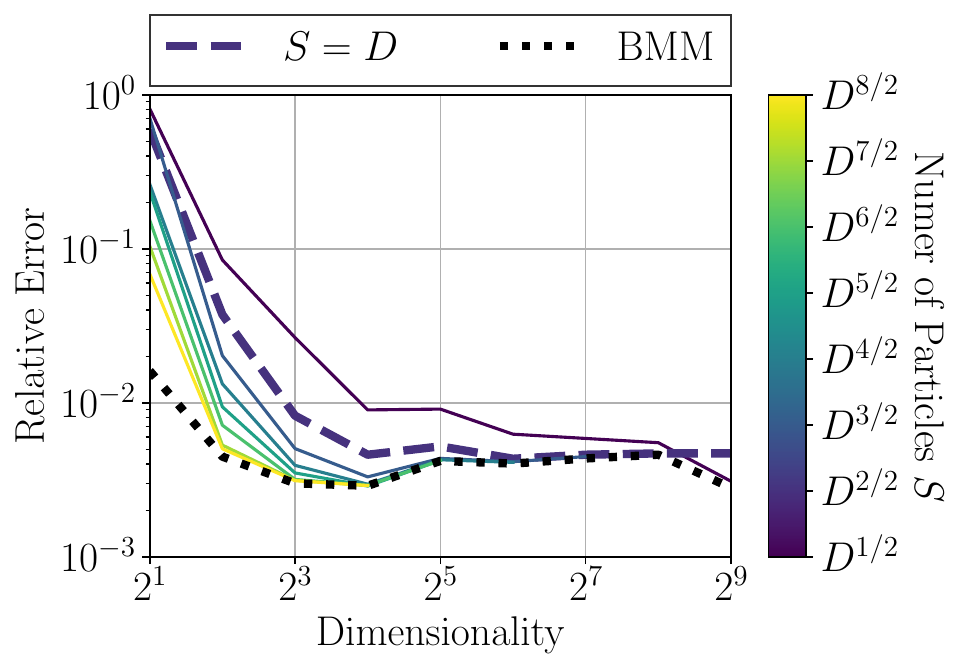}
	\caption{Relative Error}
	\label{fig:var_dim_mse}
	\end{subfigure}
	\caption{
	\textcolor{black}{Comparison of BMM to MC in terms of ECPE (left) and relative error (right) as a function of the dimensionality of $x$. 
	We calculate the relative error as 
	$\frac{1}{K}\sum_{k=1}^K||r_k-\hat{r_k}||^2/||r_k||^2$, where $r_k$ is the true mean at time step $k$ and $\hat{r}_k$ is the predicted mean.
		The dotted black line is our method BMM and the solid lines represent MC sampling. 
		The color of the solid lines indicates the number of particles $S$ as a function of the dimensionality $D$. 
		The dashed line is the sampling strategy $S=D$, which has the same computational complexity as our method $\mathcal{O}(D^2)$. 
		We show the mean over 20 runs.
	}}
\label{fig:var_dim}
\end{figure}

\textcolor{black}{
\subsection{High Dimensional Dynamics}
\label{sec:exp_high_dim}
In this experiment, we compare our inference scheme against Monte Carlo Sampling when varying the dimension $D$ and the number of Monte Carlo Samples $S$.
\subsubsection{Dataset}
We use a $D$-dimensional Ornstein-Uhlenbeck process $dx_t=(\lambda - x_t)dt + \Gamma dw_t$ for data generation. The term $\lambda \in \mathbb{R}^D$ is sampled from a normal distribution $\lambda \sim \mathcal{N}(0, I)$ and $\Gamma \in \mathbb{R}^{D\times D} $ is sampled from a Wishart distribution $\Gamma \sim \mathcal{W}( I, \mathrm{dim}( I))$. 
We vary the dimensionality on a logarithmic scale between $D=2$ and $D=512$. 
For each dimensionality we generate 100 paths with a length of $T=10s$, $\Delta t = 0.1s$. We repeat this experiment 20 times.
\
\subsubsection{Baselines}
We compare our method NSDE-BMM against NSDE-MC using a NSDE with one hidden layer in the drift neural network and a constant diffusion term.
The hidden layer size is $D$ and the drift term has dimensionality $D \times D$. We train the NSDE on one step predictions, which is a deterministically tractable
training objective (see Eq. \ref{eq:em_normal}).
After training, we use our method BMM as well as Monte Carlo sampling for testing. This enables a fair comparison between both methods as the same NSDE is used during test time.
\subsubsection{Results}
We present our main findings in Fig. \ref{fig:var_dim_ecpe}. First, we observe that, for all methods, the ECPE drops when increasing the number of dimensions.
This might be explained by the increasing complexity of the problem; the Ornstein-Uhlenbeck process is described by $O(D^2)$ many parameters, while the size of the dataset scales with $O(D)$.
Next, using a MC sampling strategy with $S=D$ particles, which has the same computational complexity as our method (see Sec.~\ref{subsec:computational_cost}), results in uncalibrated predictions regardless of the chosen dimension.
Even though Monte Carlo sampling can come close the ECPE levels of BMM, it never reaches the same ECPE level of BMM, even when using  $S=D^4$ particles. Furthermore, we frequently encounter an OOM error on a CPU and 32GB memory for NSDE-MC for high dimensions and costly sampling strategies beyond $S=D$. This can be seen by the early stopping of solid lines in Fig. \ref{fig:var_dim_ecpe} and  \ref{fig:var_dim_mse}.
}

\textcolor{black}{
Lastly, we observe in Fig. \ref{fig:var_dim_mse} that Monte Carlo sampling can give satisfactory results if the quantity of interest is the mean, i.e. not the covariance. In such cases the sampling strategy  $S=\sqrt{D}$ can give an approximation of the true mean with less than 1\% of relative error.
}

\footnotetext[5]{After correspondence with the authors, we present in this paper the correct NLL for diffWGP. In the original paper, an unknown issue caused a shift of the NLL.}

\textcolor{black}{
Our experiments confirm that fitting a NSDE becomes increasingly difficult and costly as the input dimension $D$ increases.
However, for many real-world applications, one can embed the high-dimensional input data into a lower dimensional space and infer the dynamics in this latent space\cite{Hafner2020Dream, stochastic_latent_ac}.
While this approach allows to reduce the runtime, it requires training auxiliary embedding neural networks or finite dimensional approximations to kernel mappings \cite{svrg}.
}


%% file: sections/related_work.tex
\section{Related Work}
\label{sec:related_work}

As SDEs are a common tool in many domains it is impossible to cover all related research efforts. We refer to \cite{sde_inference_overview} for a general overview of parameter inference approaches for SDEs. 
Many methods take the likelihood based approach, which necessitates estimation of the transition kernel. 
Sampling based methods together with the EM discretization on a fixed grid have been explored in \cite{brandt, petersen_95}, that  both use a simulated likelihood approach, and \cite{ll_inference_mcmc}, which proposes bridge constructions in order to create artificial observations between neighboring  observations.  
Higher order approximations on a fixed grid, such as the Milstein method, have been investigated in \cite{mcmc_milstein}. 
\cite{sde_blackbox} extends the bridge construction approach by combining it with variational inference. 
Deterministic methods with an assumed density based on the Kalman filter have been investigated in \cite{saerkkae_inference}, while \cite{gp_sde} explored a variational approximation based on Gaussian processes.
A likelihood free approach was proposed in \cite{Abbatietal19}, which instead minimizes an adversarial objective. 

A recent line of work combined SDEs with neural networks. 
\cite{nsde_raginsky} models a fully latent SDE with drift neural networks with no observations at intermediate time-steps.
Later \cite{adjoint_sde} proposed to model latent dynamical systems with neural SDEs and introduced the adjoint training method. 
Furthermore, NSDEs have been used for generative modeling of images as proposed in \cite{song2021scorebased}. 

Moment propagation through a predictor, on which our method also relies, is a well explored idea. 
Propagation of moments through a GP was explored in \cite{mm_bayesian_kernel, mm_gp} and successfully applied in a reinforcement learning scenario in \cite{pilco}. 
It was shown that a deterministic moment propagation scheme for parameter inference of Bayesian neural networks  \cite{wu2018deterministic} is superior to its sampling based counterpart. 

Our method is the first to propose a moment propagation scheme tailored towards NSDEs. We have extended  the moment propagation method through a neural network, as proposed in \cite{wu2018deterministic}, towards i) cross covariance computation by using for the first time in the context of moment propagation Stein's lemma, and ii) efficient expected Jacobian approximation of a neural network. 
To the best of our knowledge our study is the first to propose deterministic neural dynamics learning over sampling based procedures.

%% file: sections/conclusion.tex
\section{Conclusion}
We proposed a computationally efficient and deterministic approximation of the transition kernel for NSDEs. Our method enables accurate uncertainty quantification at moderate computational cost. 
We present a general-purpose methodological contribution, which is applicable beyond the use cases demonstrated in our experiments. 
Examples include vehicle behavior prediction, stock price forecasting, and system identification for robot manipulation.

While our method improves the reliability of predictive uncertainties, it does not contribute significantly to the explainability of the inferred system dynamics, which could be a risk factor in safety-critical applications. Furthermore, in potential use of our findings in fairness-sensitive applications, such as customer segmentation or crime forecasting, it should be accompanied with the recent findings of fairness research. 

Potential future extensions of our method include: \textcolor{black}{(i) learning embeddings for high-dimensional data (see also Sec.~\ref{sec:exp_high_dim})}, (ii) modeling of multiple modes by introducing auxiliary latent variables, which steer the modality and (iii) computational efficiency by  low-rank approximations of the covariance.

%% file: sections/biography.tex
\begin{IEEEbiography}[{\includegraphics[width=1in,height=1.25in,clip,keepaspectratio]{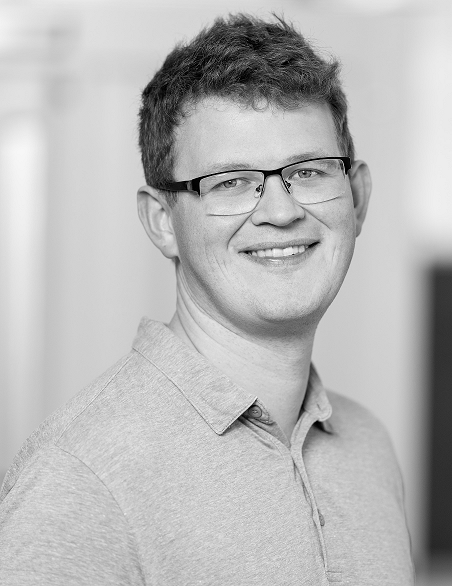}}]{Andreas Look}
 received his B.Sc. and M.Sc. from the University of Erlangen-Nuremberg in Power Engineering in 2014 and 2016, respectively. He had done research at the Institute of Fluid Mechanics and Hydraulic Machinery, University of Stuttgart, from 2016 until 2019. He joined the Bosch Center for Artificial Intelligence in 2019. His research interests are hybrid machine learning, differential equations, and optimization. \end{IEEEbiography}

\begin{IEEEbiography}[{\includegraphics[width=1in,height=1.25in,clip,keepaspectratio]{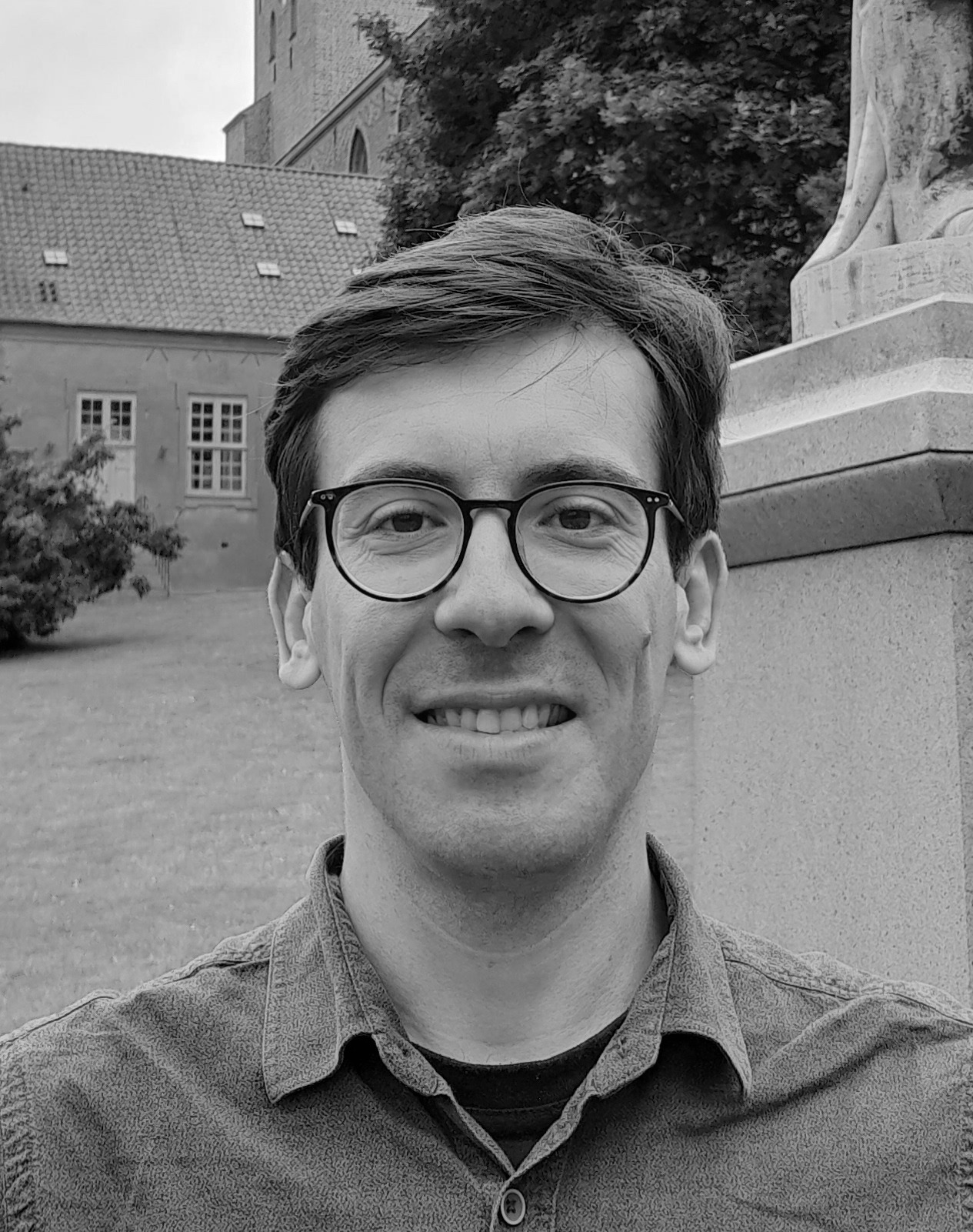}}]{Melih Kandemir} is an associate professor at the University of Southern Denmark (SDU), Department of Mathematics and Computer Science. Kandemir earned his PhD degree from Aalto University in 2013. He worked as a postdoctoral researcher at Heidelberg University, as an assistant professor at Ozyegin University in Istanbul, Turkey, and as a research group leader at Bosch Center for Artificial Intelligence. Kandemir pursues basic research on Bayesian inference and stochastic process modeling with deep neural nets with application to reinforcement learning and continual learning. Kandemir is an ELLIS member since 2021.
\end{IEEEbiography}

\begin{IEEEbiography}[{\includegraphics[width=1in,height=1.25in,clip,keepaspectratio]{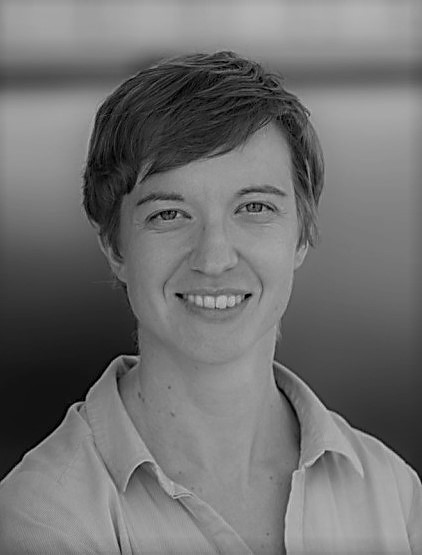}}]{Barbara Rakitsch} has been working as a research scientist at the Bosch Center for Artifical Intelligence in Renningen since 2017. Her interests lie in the area of Bayesian modeling with a focus on Gaussian processes and time-series data.
In 2014, she received her PhD in in probabilistic modeling for computational biology at the Max Planck Institute for Intelligent Systems in Tuebingen. Before joining Bosch, she worked on machine learning problems as a post-doc at the European Bioinformatics Institute in Cambridge and as a researcher in a cancer startup.
\end{IEEEbiography}

\begin{IEEEbiography}[{\includegraphics[width=1in,height=1.25in,clip,keepaspectratio]{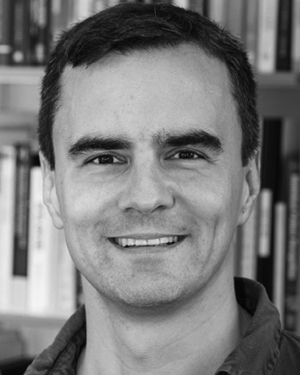}}]{Jan Peters}
is a full professor (W3) in intelligent autonomous systems, Computer Science Department, Technical University of Darmstadt, and at the same time a senior research scientist and group leader with the Max-Planck Institute for Intelligent Systems, where he heads the interdepartmental Robot Learning Group. He has received the Dick Volz Best 2007 US PhD Thesis Runner-Up Award, Robotics: Science \& Systems Early Career Spotlight, INNS Young Investigator Award,and IEEE Robotics \& Automation Society’s Early Career Award as well as numerous best paper awards. In 2015, he received an ERC Starting Grant and in 2019, he was appointed as an IEEE fellow.
\end{IEEEbiography}

%% file: main.bbl
\begin{thebibliography}{10}
\providecommand{\url}[1]{#1}
\csname url@samestyle\endcsname
\providecommand{\newblock}{\relax}
\providecommand{\bibinfo}[2]{#2}
\providecommand{\BIBentrySTDinterwordspacing}{\spaceskip=0pt\relax}
\providecommand{\BIBentryALTinterwordstretchfactor}{4}
\providecommand{\BIBentryALTinterwordspacing}{\spaceskip=\fontdimen2\font plus
\BIBentryALTinterwordstretchfactor\fontdimen3\font minus
  \fontdimen4\font\relax}
\providecommand{\BIBforeignlanguage}[2]{{%
\expandafter\ifx\csname l@#1\endcsname\relax
\typeout{** WARNING: IEEEtran.bst: No hyphenation pattern has been}%
\typeout{** loaded for the language `#1'. Using the pattern for}%
\typeout{** the default language instead.}%
\else
\language=\csname l@#1\endcsname
\fi
#2}}
\providecommand{\BIBdecl}{\relax}
\BIBdecl

\bibitem{Platt99probabilisticoutputs}
J.~C. Platt, ``{Probabilistic Outputs for Support Vector Machines and
  Comparisons to Regularized Likelihood Methods},'' in \emph{Advances in Large
  Margin Classifiers}, 1999.

\bibitem{calibration_via_aux}
J.~J. Thiagarajan, B.~Venkatesh, P.~Sattigeri, and P.~Bremer, ``{Building
  Calibrated Deep Models via Uncertainty Matching with Auxiliary Interval
  Predictors},'' in \emph{AAAI}, 2020.

\bibitem{accurate_uncertainties}
V.~Kuleshov, N.~Fenner, and S.~Ermon, ``{Accurate Uncertainties for Deep
  Learning Using Calibrated Regression},'' in \emph{ICML}, 2018.

\bibitem{calibrated_regression}
P.~Cui, W.~Hu, and J.~Zhu, ``{Calibrated Reliable Regression using Maximum Mean
  Discrepancy},'' in \emph{NeurIPS}, 2020.

\bibitem{nsde_raginsky}
B.~Tzen and M.~Raginsky, ``{Neural Stochastic Differential Equations: Deep
  Latent Gaussian Models in the Diffusion Limit},'' \emph{ArXiv}, vol.
  abs/1905.09883, 2019.

\bibitem{adjoint_sde}
X.~Li, T.~L. Wong, R.~T.~Q. Chen, and D.~Duvenaud, ``{Scalable Gradients for
  Stochastic Differential Equations},'' in \emph{AISTATS}, 2020.

\bibitem{neural_ode}
T.~Q. Chen, Y.~Rubanova, J.~Bettencourt, and D.~K. Duvenaud, ``{Neural Ordinary
  Differential Equations},'' in \emph{NeurIPS}, 2018.

\bibitem{applied_sde}
S.~S{\"a}rkk{\"a} and A.~Solin, \emph{Applied Stochastic Differential
  Equations}.\hskip 1em plus 0.5em minus 0.4em\relax Cambridge University
  Press., 2019.

\bibitem{diffgp}
P.~Hegde, M.~Heinonen, H.~L\"ahdesm\"aki, and S.~Kaski, ``{Deep learning with
  differential Gaussian process flows},'' in \emph{AISTATS}, 2019.

\bibitem{oksendal}
B.~Oksendal, \emph{{Stochastic Differential Equations: An Introduction with
  Applications}}.\hskip 1em plus 0.5em minus 0.4em\relax Springer, 1992.

\bibitem{brandt}
M.~W. Brandt and P.~Santa-Clara, ``{Simulated Likelihood Estimation of
  Diffusions with an Application to Exchange Rate Dynamics in Incomplete
  Markets},'' \emph{Journal of Financial Economics}, vol.~63, 2002.

\bibitem{petersen_95}
A.~R. Pedersen, ``{A New Approach to Maximum Likelihood Estimation for
  Stochastic Differential Equations Based on Discrete Observations},''
  \emph{Scandinavian Journal of Statistics}, vol.~22, 1995.

\bibitem{ll_inference_mcmc}
O.~Elerian, S.~Chib, and N.~Shephard, ``{Likelihood Inference for Discretely
  Observed Nonlinear Diffusions},'' \emph{Econometrica}, vol.~69, 2001.

\bibitem{HURN1999373}
A.~Hurn and K.~Lindsay, ``{Estimating the Parameters of Stochastic Differential
  Equations by Monte Carlo Methods},'' \emph{Mathematics and Computers in
  Simulation}, vol.~48, 1999.

\bibitem{Jensen18}
B.~Jensen and R.~Poulsen, ``{Transition Densities of Diffusion Processes},''
  \emph{The Journal of Derivatives}, vol.~9, 2002.

\bibitem{saerkkae_inference}
S.~S\"{a}rkk\"{a}, J.~Hartikainen, I.~S. Mbalawata, and H.~Haario, ``{Posterior
  Inference on Parameters of Stochastic Differential Equations via Non-Linear
  Gaussian Filtering and Adaptive MCMC},'' \emph{Statistics and Computing},
  vol.~25, 2015.

\bibitem{dbnn}
A.~Look and M.~Kandemir, ``{Differential Bayesian Neural Networks},'' in
  \emph{NeurIPS Workshop Bayesian Deep Learning}, 2019.

\bibitem{saerkka_smoothing}
S.~S{\"a}rkk{\"a} and J.~Sarmavuori, ``{Gaussian Filtering and Smoothing for
  Continuous-Discrete Dynamic Systems},'' \emph{Signal Processing}, vol.~93,
  2013.

\bibitem{pbp}
J.~M. Hernandez-Lobato and R.~Adams, ``{Probabilistic Backpropagation for
  Scalable Learning of Bayesian Neural Networks},'' in \emph{ICML}, 2015.

\bibitem{ghosh}
S.~Ghosh, F.~Fave, M.~Delle, and J.~Yedidia, ``{Assumed Density Filtering
  Methods for Learning Bayesian Neural Networks},'' in \emph{AAAI}, 2016.

\bibitem{wu2018deterministic}
A.~Wu, S.~Nowozin, E.~Meeds, R.~E. Turner, J.~M. Hernandez-Lobato, and A.~L.
  Gaunt, ``{Deterministic Variational Inference for Robust Bayesian Neural
  Networks},'' in \emph{ICLR}, 2019.

\bibitem{bedl}
M.~Haussmann, S.~Gerwinn, and M.~Kandemir, ``{Bayesian Evidential Deep Learning
  with PAC Regularization},'' \emph{ArXiv}, vol. abs/1906.00816, 2020.

\bibitem{steins_lemma}
J.~S. Liu, ``{Siegel's formula via Stein's identities},'' \emph{Statistics \&
  Probability Letters}, vol.~21, 1994.

\bibitem{npeet}
G.~V.~S. Shuyang~Gao and A.~Galstyan, ``{Efficient Estimation of Mutual
  Information for Strongly Dependent Variables},'' in \emph{AISTATS}, 2015.

\bibitem{aug_ode}
E.~Dupont, A.~Doucet, and Y.~W. Teh, ``{Augmented Neural ODEs},'' in
  \emph{NeurIPS}, 2019.

\bibitem{Wan00theunscented}
E.~A. Wan and R.~V.~D. Merwe, ``{The Unscented Kalman Filter for Nonlinear
  Estimation},'' in \emph{IEEE 2000 Adaptive Systems for Signal Processing,
  Communications, and Control Symposium}, 2000.

\bibitem{fastdropout}
S.~Wang and C.~Manning, ``{Fast dropout training},'' in \emph{ICML}, 2013.

\bibitem{dpmog}
L.~E. Nieto-Barajas and A.~Contreras-Cristan, ``{A Bayesian Nonparametric
  Approach for Time Series Clustering},'' \emph{Bayesian Analysis}, vol.~0,
  2014.

\bibitem{multiple_futures}
C.~Tang and R.~R. Salakhutdinov, ``{Multiple Futures Prediction},'' in
  \emph{NeurIPS}, 2019.

\bibitem{2013NDIMENSIONALCF}
\BIBentryALTinterwordspacing
M.~Bensimhou, ``{N-Dimensional cumulative function, and other useful facts
  about Gaussians and normal densities},'' Tech. Rep., 2013. [Online].
  Available:
  \url{https://upload.wikimedia.org/wikipedia/commons/a/a2/Cumulative_function_n_dimensional_Gaussians_12.2013.pdf}
\BIBentrySTDinterwordspacing

\bibitem{ece}
C.~Guo, G.~Pleiss, Y.~Sun, and K.~Q. Weinberger, ``{On Calibration of Modern
  Neural Networks},'' in \emph{ICML}, 2017.

\bibitem{dropout_gal}
Y.~Gal and Z.~Ghahramani, ``{Dropout as a Bayesian Approximation: Representing
  Model Uncertainty in Deep Learning},'' in \emph{ICML}, 2016.

\bibitem{double_vi}
H.~Salimbeni and M.~Deisenroth, ``{Doubly Stochastic Variational Inference for
  Deep Gaussian Processes},'' in \emph{NeurIPS}, 2017.

\bibitem{lindinger2020beyond}
J.~Lindinger, D.~Reeb, C.~Lippert, and B.~Rakitsch, ``{Beyond the Mean-Field:
  Structured Deep Gaussian Processes Improve the Predictive Uncertainties},''
  in \emph{NeurIPS}, 2020.

\bibitem{variational_dropout}
D.~P. Kingma, T.~Salimans, and M.~Welling, ``{Variational Dropout and the Local
  Reparameterization Trick},'' in \emph{NeurIPS}, 2015.

\bibitem{kandemir2022evidential}
M.~Kandemir, A.~Akg{\"u}l, M.~Haussmann, and G.~Unal, ``{Evidential Turing
  Processes},'' in \emph{ICLR}, 2022.

\bibitem{Abbatietal19}
G.~Abbati, P.~Wenk, M.~Osborne, A.~Krause, B.~Sch{\"o}lkopf, and S.~Bauer,
  ``{{AR}e{S} and {M}a{RS} Adversarial and {MMD}-Minimizing Regression for
  {SDE}s},'' in \emph{ICML}, 2019.

\bibitem{air_quality}
S.~Zhang, G.~Bin, D.~Anlan, H.~Jing, X.~Ziping, and C.~S. Xi, ``{Cautionary
  tales on air-quality improvement in Beijing},'' \emph{Proceedings of the
  Royal Society: Mathematical,Physical and Engineering Sciences}, vol. 473,
  2017.

\bibitem{wishart_sde}
M.~Jorgensen, M.~P. Deisenroth, and H.~Salimbeni, ``{Stochastic Differential
  Equations with Variational Wishart Diffusions},'' in \emph{ICML}, 2020.

\bibitem{robot_arm}
D.~Agudelo-España, A.~Zadaianchuk, P.~Wenk, A.~Garg, J.~Akpo, F.~Grimminger,
  J.~Viereck, M.~Naveau, L.~Righetti, G.~Martius, A.~Krause, B.~Sch{\"o}lkopf,
  S.~Bauer, and M.~W{\"u}thrich, ``{A Real-Robot Dataset for Assessing
  Transferability of Learned Dynamics Models },'' in \emph{ICRA}, 2020.

\bibitem{Hafner2020Dream}
D.~Hafner, T.~Lillicrap, J.~Ba, and M.~Norouzi, ``{Dream to Control: Learning
  Behaviors by Latent Imagination},'' in \emph{ICLR}, 2020.

\bibitem{stochastic_latent_ac}
A.~X. Lee, A.~Nagabandi, P.~Abbeel, and S.~Levine, ``{Stochastic Latent
  Actor-Critic: Deep Reinforcement Learning with a Latent Variable Model},'' in
  \emph{NeurIPS}, 2020.

\bibitem{svrg}
M.~Alioscha{-}P{\'{e}}rez, M.~C. Oveneke, and H.~Sahli, ``{SVRG-MKL: A Fast and
  Scalable Multiple Kernel Learning Solution for Features Combination in
  Multi-Class Classification Problems},'' \emph{{IEEE} Trans. Neural Networks
  Learn. Syst.}, vol.~31, no.~5, 2020.

\bibitem{sde_inference_overview}
J.~Jeisman, ``{Estimation of the Parameters of Stochastic Differential
  Equations},'' Ph.D. dissertation, Queensland University of Technology, 2005.

\bibitem{mcmc_milstein}
S.~Pieschner and C.~Fuchs, ``{Bayesian Inference for Diffusion Processes: Using
  Higher-Order Approximations for Transition Densities},'' \emph{Royal Society
  Open Science}, vol.~7, 2020.

\bibitem{sde_blackbox}
T.~Ryder, A.~Golightly, A.~S. McGough, and D.~Prangle, ``{Black-Box Variational
  Inference for Stochastic Differential Equations},'' in \emph{ICML}, 2018.

\bibitem{gp_sde}
C.~Archambeau, D.~Cornford, M.~Opper, and J.~S.-Taylor, ``{Gaussian Process
  Approximations of Stochastic Differential Equations},'' \emph{Proceedings of
  Machine Learning Research}, vol.~1, 2007.

\bibitem{song2021scorebased}
Y.~Song, J.~Sohl-Dickstein, D.~P. Kingma, A.~Kumar, S.~Ermon, and B.~Poole,
  ``{Score-Based Generative Modeling through Stochastic Differential
  Equations},'' in \emph{ICLR}, 2021.

\bibitem{mm_bayesian_kernel}
J.~Candela, A.~Girard, J.~Larsen, and C.~Rasmussen, ``{Propagation of
  uncertainty in Bayesian kernel models - application to multiple-step ahead
  forecasting},'' in \emph{ICASSP}, 2003.

\bibitem{mm_gp}
M.~P. Deisenroth, M.~F. Huber, and U.~D. Hanebeck, ``{Analytic Moment-Based
  Gaussian Process Filtering},'' in \emph{ICML}, 2009.

\bibitem{pilco}
M.~Deisenroth and C.~Rasmussen, ``{PILCO: A Model-Based and Data-Efficient
  Approach to Policy Search},'' in \emph{ICML}, 2011.

\end{thebibliography}
